\documentclass[11pt,english]{article}
\usepackage[T1]{fontenc}
\usepackage{lmodern}
\usepackage[utf8]{inputenc}
\usepackage[margin=1in]{geometry}
\usepackage{abstract}
\usepackage{array}
\usepackage{booktabs}
\usepackage{tablefootnote}
\usepackage{amsfonts}
\usepackage{graphicx}
\usepackage[font={footnotesize,it}]{caption}
\usepackage{multirow}

\usepackage[usenames,dvipsnames]{xcolor}
\usepackage[algo2e,ruled]{algorithm2e}
\usepackage{amssymb}
\usepackage{amsmath}
\usepackage{amsthm}
\usepackage{cite}
\usepackage{url}

\usepackage{enumitem}
\setlist{nosep,leftmargin=3em}

\makeatletter

\usepackage{cleveref}

\Crefname{algocf}{Algorithm}{Algorithms}
\Crefname{ALC@unique}{Line}{Lines}
\crefname{line}{line}{lines}

\usepackage{babel}
\providecommand{\corollaryname}{Corollary}
\providecommand{\lemmaname}{Lemma}
\providecommand{\propositionname}{Proposition}
\providecommand{\theoremname}{Theorem}

\theoremstyle{plain}

\theoremstyle{plain}
\newtheorem{theorem}{\protect\theoremname}[section]
\newtheorem{lemma}[theorem]{\protect\lemmaname}
\newtheorem*{proposition*}{\protect\propositionname}
\newtheorem{proposition}[theorem]{\protect\propositionname}
\theoremstyle{plain}
\newtheorem{corollary}[theorem]{\protect\corollaryname}
\newtheorem*{theorem*}{Theorem}

\begin{document}
\title{Privacy of the last iterate in cyclically-traversed DP-SGD on nonconvex composite losses}
\author{
Weiwei Kong\footnote{Google Research, Email: {\tt{weiweikong@google.com}}, ORCID: {\tt{0000-0002-4700-619X}}}
\and
M\'onica Ribero\footnote{Google Research, Email: \tt{mribero@google.com}}
}

\maketitle
\begin{abstract}

Differentially-private stochastic gradient descent (DP-SGD) is a family of iterative machine learning training algorithms that privatize gradients to generate a sequence of differentially-private (DP) model parameters. It is also the standard tool used to train DP models in practice, even though most users are only interested in protecting the privacy of the final model. 
Tight DP accounting for the last iterate would minimize the amount of noise required while maintaining the same privacy guarantee and potentially increasing model utility. However, last-iterate accounting is challenging, and existing works require strong assumptions not satisfied by most implementations.
These include assuming (i) the global sensitivity constant is known --- to avoid gradient clipping; (ii) the loss function is Lipschitz or convex; and (iii) input batches are sampled randomly.

In this work, we forego any unrealistic assumptions and provide privacy bounds for the most commonly used variant of DP-SGD, in which data is traversed cyclically, gradients are clipped, and only the last model is released. More specifically, we establish new R\'enyi differential privacy (RDP) upper bounds for the last iterate under realistic assumptions of small stepsize and Lipschitz smoothness of the loss function. Our general bounds also recover the special-case convex bounds when the weak-convexity parameter of the objective function approaches zero and no clipping is performed. The approach itself leverages optimal transport techniques for last iterate bounds, which is a nontrivial task when the data is traversed cyclically and the loss function is nonconvex. 
\end{abstract}

\global\long\def\rn{\mathbb{R}^{n}}%
\global\long\def\R{\mathbb{R}}%
\global\long\def\r{\mathbb{R}}%
\global\long\def\n{\mathbb{N}}%
\global\long\def\c{\mathbb{C}}%
\global\long\def\pt{\mathbb{\partial}}%
\global\long\def\lam{\lambda}%
\global\long\def\argmin{\operatorname*{argmin}}%
\global\long\def\argmax{\operatorname*{argmax}}%
\global\long\def\esssup{\operatorname*{ess\ sup}}%
\global\long\def\dom{\operatorname*{dom}}%
\global\long\def\prox{\operatorname{prox}}%
\global\long\def\cConv{\overline{{\rm Conv}}\,}%

\section{Introduction}

Differential privacy (DP) is an approach to capture the sensitivity of an algorithm to any individual user's data  and is frequently used in both industrial and government applications (see the book by \cite{DR14} for a rich introduction).  Given a possibly nonprivate computation $f$, a desired level of DP (or privacy budget) $\varepsilon$ is generally achieved by bounding the global sensitivity\footnote{The maximum change in the mechanism's output caused by changes in a single user or data point.} of $f$ and then adding noise to its output. This noise is typically calibrated to the sensitivity and $\varepsilon$ in order to obscure the contributions of a single input example. Conversely, given a mechanism $\cal A$ for making a computation differentially private, a method for determining the level of DP obtained by $\cal A$ is often called a DP accounting method. 

Differentially-private stochastic gradient descent (DP-SGD) refers to a family of popular first-order methods for training model weights with DP\cite{CMS11, SCS13, BST14, ACGM+16}.
At a high level, a DP-SGD method first computes the gradients of a given set of per-example loss functions with respect to the model weights and applies an algorithm ${\cal A}$ to obtain a private gradient $\cal G$.
The private gradient $\cal G$ is then used in some first-order optimization scheme, e.g., SGD, Adam, or AdaGrad, to update model weights. More precisely, ${\cal A}$ consists of (i) scaling the per-example loss gradients (a.k.a. gradient clipping) to reduce sensitivity, (ii) adding independent and identically distributed (i.i.d.) Gaussian noise $\cal Z$ to each of the scaled gradients, and (iii) summing the noised gradients to obtain $\cal G$. 
In general, the higher the variance of $\cal Z$ is, the lower the utility of the final trained model.

Depending on the optimization scheme, and the assumptions on how the user-level loss functions are obtained,
existing DP accounting methods for DP-SGD can differ significantly. For example, when only the last iterate of DP-SGD is released, 
existing accounting methods require both sophisticated machinery and numerous strong assumptions to provide tight DP bounds.
Some of these strong assumptions, that almost never hold in practice, include (i) the input data is sampled randomly at each DP-SGD iteration, (ii) the loss functions are convex, (iii) the global DP-SGD sensitivity is known beforehand, and (iv) the intermediate model weights  are bounded.

This work develops tighter privacy analyses for last-iterate DP-SGD under more realistic settings than existing works. Consequently, our analyses enable implementations of DP-SGD that apply Gaussian noise $\cal Z$ with \textit{lower variance} than existing work,  and as a consequence, obtain higher utility at the same privacy budget $\varepsilon$. 
More specifically, we develop a family of R\'enyi DP (RDP) bounds on the last iterate of DP-SGD, which are novel in that they:
\begin{itemize}
    \item[(i)]do not assume knowledge of the global sensitivity constant and, hence, are valid with or without gradient clipping;
    \item[(ii)] hold for both the nonconvex and convex settings under significantly fewer assumptions than other works;
    \item[(iii)]are parameterized by a weak convexity\footnote{A function $f$ is $m$-weakly-convex if $f+m\|\cdot\|^2/2$ is convex.} parameter $m\geq 0$, for which one of the bounds smoothly converges to a similar one in the convex setting as $m\to0$.
\end{itemize}

\subsection{Background}

We begin by formally stating the problem of interest, describing common terminology and notation, and specifying the DP-SGD variant under consideration. We then briefly describe the Privacy Amplification by Iteration (PABI) argument of \cite{feldman2018privacy} and discuss the difficulties of generalizing this argument to more practical settings. 

\vspace*{1em}
\noindent\textbf{Problem of interest}.
We develop RDP bounds for the last iterate
of a DP-SGD variant applied to the nonsmooth composite optimization problem
\begin{equation}
\min_{x\in\rn}\left\{\phi(x) := \frac{1}{k}\sum_{i=1}^{k}f_{i}(x)+h(x)\right\} \label{eq:main_prb}
\end{equation}
where  $h$ is convex and proper lower-semicontinuous and $f_{i}$ is continuously differentiable on the domain of $h$. 
Notice that the assumption on $h$ encapsulates (i) common nonsmooth regularization functions such as the $\ell_1$-norm $\|\cdot\|_1$, nuclear matrix norm $\|\cdot\|_*$, elastic net regularizer and (ii) indicator functions on possibly unbounded closed convex sets. A common setting in practice is when $(1/k)\sum_{i=1}^k f_i(x)$ corresponds to a softmax cross-entropy loss function and $h(x)$ corresponds to an $\ell_1$- or $\ell_2$-regularization function.

\vspace*{1em}
\noindent\textbf{Common terminology}.  An input data collection $X=\{x_i\}_{i=1}^k$ is said to be \textit{traversed cyclically} (or cyclically-traversed) in batches $\{B_t\}$ of size $b$ if $B_t$ contains $\{x_{b(t-1)+1},\ldots, x_{bt}\}$ for the first $t\leq k/b$ batches\footnote{For simplicity, we assume $b$ divides $k$ throughout.}, and the the rest of the batches cycle between $B_1, \ldots, B_{k/b}$ in order. \textit{Cyclically-traversed DP-SGD} is a variant of DP-SGD where the input data is traversed cyclically\footnote{Cyclically traversed is also known in the literature as incremental gradient \cite{LZ24}.}. A \textit{dataset pass} occurs when the input data (e.g., $X$ above) in a cyclically-traversed DP-SGD run has been used, and the next batch of inputs is the same as the first batch of inputs at the beginning of the dataset pass. \textit{Gradient clipping} is the process of orthogonally projecting a gradient vector in $\rn$ to a Euclidean ball of radius $C$ centered at the origin. The parameter $C$ is typically called the $\ell_2$-norm clip value.
In this work, we say a function is a \textit{randomized operator} if it consists of applying some deterministic operator to an input and adding random noise to resulting output. An operator ${\cal T}$
is said to be $L$-\textit{Lipschitz} if $\|{\cal T}(x)-{\cal T}(y)\|\leq L\|x-y\|$
for every $x$ and $y$, and ${\cal T}$ is said to be \textit{nonexpansive} if
it is 1-Lipschitz.

\vspace*{1em}
\noindent\textbf{Common notation}.
Let $[n]=\{1,\ldots,n\}$ for any positive integer $n$. Let $\r$ denote the set of real numbers and $\rn=\r\times\cdots\times\r$ denote the $n$-fold Cartesian product of $\r$. Let $(\langle \cdot, \cdot \rangle, \rn)$ denote a Euclidean space over $\rn$ and denote $\|\cdot\| :
= \sqrt{\langle \cdot, \cdot \rangle}$ to be the induced norm. The \textit{domain} of a function $\phi:\rn\mapsto(-\infty,\infty]$
is $\dom\phi:=\{z\in\rn:\phi(z)<\infty\}$. The function $\phi$ is
said to be \textit{proper} if $\dom\phi\neq\emptyset$. A function $\phi:\rn\mapsto(-\infty,\infty]$
is said to be \textit{lower semicontinuous} if $\liminf_{x\to x_{0}}\phi(x)\geq\phi(x_{0})$.
The set of proper, lower semicontinuous, convex functions over $\rn$
is denoted by $\cConv(\rn)$. 
The \textit{clipping operator} is given by
\begin{align}
\label{eq:ClipC_def}
{\rm Clip}_{C}(y) := y \cdot \min\left\{1, \frac{C}{\|y\|}\right\},
\end{align}
and the \textit{proximal operator} for a proper convex function $\psi$ is defined as 
\begin{equation}
\prox_{\psi}(z_{0})=\argmin_{z\in\rn}\left\{ \psi(z)+\frac{1}{2}\|z-z_{0}\|^{2}\right\} \quad\forall z_{0}\in\rn. \label{eq:prox_def}
\end{equation}
It is well-known that ${\rm prox}_{\psi}(\cdot)$ is nonexpansive
(see, for example \cite[Theorem 6.42]{beck2017first}) and that ${\rm Clip}_{C}(y)$ is the proximal operator for the (convex) indicator function of the set $\{x: \|x\|\leq C\}$. 

The $\infty$-Wasserstein metric ${\cal W}_{\infty}(\mu,\nu)$
is the smallest real number $w$ such that for $X\sim\mu$ and $Y\sim\nu$,
there is a joint distribution on $(X,Y)$ where $\|X-Y\|\leq w$ almost
surely, i.e., ${\cal W}_{\infty}(\mu,\nu)=\inf_{\gamma\sim\Gamma(\mu,\nu)}\esssup_{(x,y)\sim\gamma}\|x-y\|$,
where $\Gamma(\mu,\nu)$ is the collection of measures on $\rn\times\rn$
with first and second marginals $\mu$ and $\nu$, respectively. For any probability distributions $\mu$ and $\nu$ with $\nu \ll \mu$, the
\textit{R\'enyi divergence} of order $\alpha\in(1,\infty)$ is 
\begin{equation}
D_{\alpha}(\mu\|\nu)=\frac{1}{\alpha-1}\log\int\left[\frac{\mu(x)}{\nu(x)}\right]^{\alpha}\nu(x)\,dx, \label{eq:renyi_div_def}
\end{equation}
where we take the convention that $0/0=0$. For $\nu \not \ll \mu$, we define $D_{\alpha}(\mu\|\nu)=\infty$. For
parameters $\tau\geq0$ and $\alpha\geq1$, the \textit{shifted R\'enyi
divergence} is 
\begin{equation}
D_{\alpha}^{(\tau)}(\mu\|\nu):=\inf_{\gamma}\left\{ D_{\alpha}(\gamma\|\nu):{\cal W}_{\infty}(\mu,\gamma)\leq\tau\right\} \label{eq:shifted_div}
\end{equation}
for any probability distributions $\mu$ and $\nu$ over $\rn$. Given random variables $X\sim\mu$ and $Y\sim\nu$, we denote $D_{\alpha}(X\|Y)=D_{\alpha}(\mu\|\nu)$ and $D_{\alpha}^{(\tau)}(X\|Y)=D_{\alpha}^{(\tau)}(\mu\|\nu)$.

We consider the swap model for differential privacy. We say two datasets $S$ and $S'$ are neighbors, denoted as $S\sim S'$, if $S'$ can be obtained by swapping one record.
A randomized algorithm $\cal A$ is said to be $(\alpha, \varepsilon)$-RDP if, for every pair of neighboring datasets $S$ and $S'$ in the domain of $\cal A$, we have $D_\alpha({\cal A}(S)\|{\cal A}(S'))\leq \varepsilon$. 

${\cal A}$ satisfies local DP if for all records $x_i$ and $x_j$,   $D_\alpha({\cal A}(x_i)\|{\cal A}(x_j))\leq \varepsilon$.
Finally, we use the following variable conventions: $\ell$ is the number of batches (or iterations) in a dataset pass, $E$ is the number of dataset passes, $T = E \cdot \ell$ is the total number of iterations, $k$ is the total number of per-example losses, $b$ is the batch size, $\lam$ is the DP-SGD stepsize, and $C$ is the clipping norm.

\vspace*{1em}
\noindent\textbf{DP-SGD variant}. 
 Algorithm~\ref{alg:dp_sgd} outlines the specific variant of DP-SGD applied to \eqref{eq:main_prb}. This variant takes as input $k$ per-example loss functions $\{f_i \}^k_{i=1}$, the number of iterations $T$, iid samples $\{N_t\}^T_{t=1}$ from a spherical Gaussian distribution ${\cal N}(0, \sigma^2I)$, initial model weights $X_0$, batch size, stepsize, and $\ell_2$-clipping norm values. Model weights are updated as follows. At time step $t$, the algorithm (i) selects a batch of examples by cyclically traversing $\{f_i \}^k_{i=1}$, (ii) computes the average $g_t$ of clipped per example gradients at $X_{t-1}$, and (iii) updates $X_{t-1}$ using a noisy gradient. Finally, the algorithm returns the last iterate, $X_T$.

\DontPrintSemicolon 
\begin{algorithm2e}[ht]
\caption{Cyclically-traversed last-iterate DP-SGD}
\label{alg:dp_sgd}
\KwIn{$\{f_{i}\}_{i=1}^{k}$, $h$, iid samples $\{N_{t}\}\subseteq\rn$ from ${\cal N}(0,\sigma^{2}I)$, $X_{0}\in\dom h$;}
\KwData{batch size $b$, stepsize $\lam$, clipping norm $C$, iteration limit $T$, steps per dataset pass $\ell$; }
\KwOut{$X_{T}\in\dom h$;}
\vspace*{0.3em}
\For{$t=1,\ldots,T$}
{
$j_t \gets b(t-1) \bmod k$\;
$B_{t}\gets\{j_t + 1, \ldots, j_t + b\}$\;
$g_{t}\gets(1/b)\sum_{i\in B_{t}}{\rm Clip}_{C}(\nabla f_{i}(X_{t-1}))$\;
$X_{t}\gets\prox_{\lam h}\left(X_{t-1}-\lam g_{t}+N_{t}\right)$
}
\KwRet{$X_{T}$}
\end{algorithm2e}

\subsection{Outline of approach}

We now outline our approach of tackling the problem of interest. A more formal treatment is given in Section~\ref{sec:privacy_bds}.

To motivate our approach, we provide a brief overview of previous well-known methods. An early approach of analyzing Algorithm~\ref{alg:dp_sgd} is to develop a bound based on local DP for a single dataset pass and extend this bound for multiple passes (see, for example, \cite{SCS13, CGK24}). While straightforward, this approach can be overly restrictive in a centralized setting. Privacy Amplification by Subsampling (PABS) (see subsequent work  \cite{CGKK+24} for a comparison of different sampling methods)  improves on the previous approach in certain regimes. Although this method allows for clean privacy accounting, its reliance on Poisson subsampling makes it impractical for large-scale applications (see Appendix~\ref{app:random_batch_sampling} for an extended discussion). The work started by \cite{feldman2018privacy} addressed the limitations of PABS with Privacy Amplification by Iteration (PABI), which achieves a bound for releasing the final DP-SGD iterate. This PABI bound improves the baseline bound in certain regimes incorporating a contraction factor dependent on the loss function's convexity parameters.

Our approach is inspired by PABI, but relaxes several of its convexity assumptions. For added context, we briefly review PABI  below.
Under the assumption that the loss function of \eqref{eq:main_prb} is convex and $Q$-Lipschitz, and that $h$ is the indicator of a closed convex set, \cite{feldman2018privacy} shows that the DP-SGD update in Algorithm~\ref{alg:dp_sgd} with small constant stepsize $\lam$ \textit{and no gradient clipping} is a nonexpansive operator. This property can be then combined with the following technical result about nonexpansive operators.

\begin{theorem}
Suppose we are given iterates $\{X_t\}$ and $\{X_t'\}$, nonexpansive operators $\{\phi_t\}$ and $\{\phi_t'\}$, iid Gaussian random variables $\{N_t\}$, and scalars $\{s_t\}$ satisfying
\[
X_t = \phi_t(X_{t-1}) + N_t,\quad X_t' = \phi_t'(X_{t-1}') + N_t, \quad \sup_{x}\|\phi_t(x) - \phi_t'(x')\| \leq s_t \quad \forall t\in[T].
\]
For any scalar sequences $\{a_t\}$ and $\{z_t\}$ satisfying
\begin{equation}
z_t = \sum_{i\leq t} s_i - \sum_{i\leq t} a_i \geq 0, \quad z_t \geq 0, \quad a_t \geq 0, \quad \forall t\in[T],
\label{eq:zt_at_bd}
\end{equation}
we obtain the following last-iterate shifted R\'enyi divergence bound:
\begin{equation}
D_\alpha^{(z_T)}(X_T\|X_T') - D_\alpha(X_0\|X_0') \leq \frac{\alpha}{2\sigma^2} \sum_{t=1}^T a_t^2 =: {\cal R}_T(\{a_t\}) \quad \forall T\geq 1,\quad \forall \alpha \geq 1. \label{eq:pabi_bd}
\end{equation}
\end{theorem}
More specifically, assuming that the DP-SGD iterates first differ at index $t^*$, i.e., $X_{t^*}' \neq X_{t^*}$ and $X_{t}' = X_{t}$ 
for every $t < t^*$, the operators $\{\phi_t\}$ and $\{\phi_t'\}$ in the above theorem can be formed with $s_{t^*} = 2\lam Q$ and $s_{t} = 0$ for every $t \neq t^*$. Consequently, one can select $\{a_t\}$ so that the shift satisfies $z_T = 0$ and obtain a closed form bound ${\cal R}_T(\{a_t\})= \Theta(\alpha [\lam Q / \sigma]^2)$ in \eqref{eq:pabi_bd}, which yields a corresponding RDP bound when $X_0 = X_0'$. The generalization to multiple dataset passes follows similarly but  the final bound scales with the number of dataset passes $E$. Specifically, we have ${\cal R}_T(\{a_t\})= \Theta(\alpha [E / \ell] \cdot [\lam Q / \sigma]^2)$.

In the more practical setting where (i) the loss function in \eqref{eq:main_prb} is nonconvex and \textit{not} necessarily $Q$-Lipschitz, (ii) gradient clipping \textit{is} applied, and (iii) $h$ is nonsmooth, it is no longer clear to what extent the corresponding DP-SGD operators $\{\phi_t\}$ and $\{\phi_t'\}$ are nonexpansive or how $\{s_t\}$ should be obtained. Furthermore, the first inequality of \eqref{eq:zt_at_bd} no longer holds, and additional technical issues arise when analyzing the case of multiple dataset passes. 

\vspace*{1em}
\noindent\textbf{Our approach}.
We generalize the above argument and combine it with additional analyses of weakly-convex functions and proximal operators to relax several strong assumptions. A sketch of our approach is given below, and formal arguments can be found in subsequent sections.

\vspace*{1em}
\noindent\textit{General operator analysis}. In Lemma~\ref{lem:main_recurrence}, we study properties of operators $\phi$ and $\phi'$ satisfying
\begin{equation}
\sup_{u}\|\phi'(u)-\phi(u)\|\leq s,\quad \|\Phi(x) - \Phi(y)\|\leq L\|x-y\| + \zeta, \quad \forall \Phi\in\{\phi,\phi'\}.
\label{eq:phi_reg}
\end{equation}
Note that if $\Phi$ is H\"older continuous, then it can be shown \cite{liang2024unified} that it satisfies the second inequality in \eqref{eq:phi_reg}\footnote{More specifically, if $\Phi$ is  $\eta$-H\"older continuous with modulus $H$, then \eqref{eq:phi_reg} holds with $L = H\rho / 2$ and $\zeta = H\eta([1-\eta]/\rho)^{(1-\eta)/\eta}$ for any $\rho > 0$.}. Using these properties, we then establish in Proposition~\ref{prop:tech_div} that if $\{Y_t\}$ and $\{Y_t'\}$ are generated by a specific sequence of randomized proximal operators using $\phi$ and $\phi'$, respectively, then (roughly)
\[
D_\alpha(Y_T\|Y_T') - D_\alpha(Y_0\|Y_0') \preceq \frac{\alpha}{\sigma^2} \cdot {\cal{F}}(T), 
\]
where ${\cal F}:\mathbb{R} \mapsto \mathbb{R}$ is non-decreasing and dependent on some assumptions on $\{Y_t\}$ and $\{Y_t'\}$. 
% Note that these analyses generalize similar arguments in \cite{feldman2018privacy, altschuler2022privacy} for nonexpansive operators\footnote{See the \textit{Technical Remarks} portion of the next subsection for why this  generalization is nontrivial.}.
More specifically, we obtain a sequence of parameterized shifted R\'enyi divergence bounds similar to \eqref{eq:pabi_bd}, 
while dealing with the challenge of nonconvexity. 
 In the setting of one dataset pass, we derive the bound by solving a related quadratic programming problem on a similar set of residuals $\{ a_t\}$ as in \eqref{eq:zt_at_bd} (see Appendix~\ref{app:residuals} for details). 

\vspace*{1em}
\noindent\textit{Lipschitz properties of the DP-SGD update}. 
Denoting ${\cal A}_\lam(\cdot)$ as the DP-SGD update function\footnote{Or any SGD-like update as in \eqref{eq:gen_dp_sgd_update}.}, i.e., $X_t = {\cal A}_\lam(X_{t-1})$ for every $t$ in Algorithm~\ref{alg:dp_sgd}, we show in Proposition~\ref{prop:prox_lipsch} that --- depending on our assumptions on $h$ and stepsize $\lam$ --- the operator ${\cal A}_\lam(\cdot)$ satisfies the second inequality of \eqref{eq:phi_reg} with $\Phi=A_\lam$ for different values of $L$ and $\zeta$. 

More specifically, when the domain of $h$ is bounded, we have $(L, \zeta)=(1,2\lam C)$ in \eqref{eq:phi_reg} for clipping norm $C$. On the other hand, when $\lam$ is sufficiently small, we have $\zeta = 0$ and $L$ being a constant that (i) tends to $\sqrt{2}$ when the weak convexity parameter $m$ tends to zero, i.e., $f$ becomes more convex; and (ii) tends to one when no clipping is performed and $m$ in (i) tends to zero. 
This continuity with respect to the weak convexity parameter appears to be new, and it is proved in Appendix~\ref{app:sgd_analysis} by using topological properties about weakly-convex functions and proximal operators.

\vspace*{1em}
\noindent\textit{Privacy bounds for DP-SGD}. For neighboring DP-SGD iterates $X_T$ and $X_T'$ , we combine the above results in Theorems~\ref{thm:dp_sgd_bd1} and \ref{thm:general_compl} to obtain RDP bounds of the form
\begin{equation}
D_{\alpha}(X_{T}\|X_{T}') \preceq \frac{\alpha}{\sigma^2} \cdot {\cal{B}}_\lam(C, b, T, \ell),\label{eq:intro_bd}
\end{equation}
where $C$, $b$, $T$, $\ell$, are as in Algorithm~\ref{alg:dp_sgd}. More specifically, 
assuming that the DP-SGD iterates are contained within an $\ell_2$ ball of diameter $d_h$ and each $\nabla f_i$ is Lipschitz continuous, we obtain 
\eqref{eq:intro_bd} with $B_\lam(C,b,T,\ell)=(L_\lam d_h + \lam C / b)^2$ for for some $L_\lam = \sqrt{1+\kappa \lam m}$, $\kappa \leq 4$, and small enough $\lam$. When the iterates are (possibly) unbounded, we obtain \eqref{eq:intro_bd} with 
\begin{equation}
B_\lam(C,b,T,\ell)=\left(1+\left[\frac{T}{\ell}\right]\left[\frac{ L^{2\ell}_\lam}{\sum_{i=1}^\ell L^{2i}_\lam}\right]\right)\left(\frac{\lam C}{b}\right)^2. \label{eq:smooth_b}
\end{equation}

\subsection{Related work}
\label{sec:lit_review}

We first present high-level descriptions of related works in the convex and nonconvex settings, followed by more general works that use advanced composition to obtain loose bounds on the last iterate.  We then conclude with some summary tables and figures, and a discussion of technical nuances that carefully compares our work to existing literature.

\vspace*{1em}
\noindent\textbf{Convex setting}.
Given the challenge of proving tight bounds in the general setting, a number of prior analyses focus on the convex case. 
Works by \cite{feldman2018privacy, feldman2020private} additionally assume  Lispchitz continuity of the loss function to obtain results. The work of \cite{CYS21} studies multiple passes over the data, but their results only apply to the smooth, strongly convex, and full batch setting without clipping. The work of \cite{YS22} improves these results and extend them to mini-batches both with ``shuffle and partition'' and ``sampling without replacement'' strategies. 
Similarly, results in by \cite{altschuler2022privacy}, and its extension \cite{ABT24}, consider only convex Lipschitz smooth functions. The contemporary work of \cite{BSA24} introduces the shifted interpolated process under $f$-DP, allowing for tight characterizations of privacy by iteration for R\'enyi and other generalized DP definitions.

Notice that none of the above works study clipping, all assume access to the Lipschitz constant of the loss function, and require convexity, limiting their practical viability.

\vspace*{1em}
\noindent\textbf{Nonconvex setting}. We now discuss papers that do not require convexity of the loss function.
\cite{asoodeh2023privacy} analyze the privacy loss dynamics for nonconvex functions, but their analysis differs from ours in two ways. First, they assume that their DP-SGD batches are obtained by Poisson sampling or sampling without replacement. Second, their results require numerically solving a minimization problem that can be hard in practice.

A contemporary work\footnote{Appearing after the first version of this preprint.} by  \cite{CL24} derives bounds under the assumption that the loss gradient is H\"older continuous and the loss function is Lispchitz continuous when it is also convex. However, this work needs an additional assumption, that constants $L > 0$ and $\eta \in (0,1]$ satisfying $\|\nabla f(x) - \nabla f(y)\| \leq L\| x-y\|^{\eta}$ are known and used in a specific optimization subproblem.
While \cite{CL24} focuses on tight theoretical bounds under specific conditions (such as the full batch and single-epoch setting), we prioritize bridging the gap between theory and practice by addressing the complexities of real-world deployments. 

% Our work, therefore, provides a more practical and applicable analysis, encompassing crucial elements like clipping, shuffling, and non-convex loss functions. 
% In contrast, we use a more general and less conservative approach with an additive constraint (see \cref{eq:phi_reg}). While this may lead to looser bounds in idealized settings (e.g., full-batch gradient descent with convex losses where the H\"older condition is easily verifiable), our approach offers significant advantages in realistic scenarios. Specifically,the work by \cite{CL24} relies on solving an optimization problem within their bound, rendering it impractical when incorporating cyclic or shuffled batching strategies, a key component of real-world DP-SGD.

\vspace*{1em}
\noindent\textbf{Non-specialized analyses}.
Prior works on DP-SGD accounting often rely on loose bounds that allow for the release of intermediate updates  \cite{kaissis2021unified, BST14, ACGM+16}. These works rely on differential privacy advanced composition results \cite{kairouz2015composition}, resulting in noise with standard deviation that scales as $\sqrt{T}$ \cite{BST14, ACGM+16}. Alternatively, using disjoint batches decreases the dependence from the number of iterations $T$ to the number of epochs $E$ (see the first row in Table~\ref{tab:bd_summary}). However, the assumption that all intermediate updates are released can be stringent for certain loss regimes \cite{feldman2018privacy, feldman2020private}, where this bound can be contracted based on loss smoothness and convexity parameters, if only the last iterate is released.

While our work focuses on the privacy guarantees of DP-SGD, it's important to acknowledge the parallel research efforts exploring the convergence properties of shuffling methods. Recent studies, such as \cite{LZ24} and \cite{CD24}, have established convergence bounds for strongly convex and/or smooth functions in various settings, albeit without considering privacy. These works provide valuable insights into the optimization behavior of shuffling techniques, which can inform future research at the intersection of privacy and optimization.

\begin{figure}[!ht]
\begin{centering}
\hspace*{\fill}
\includegraphics[width=0.4\columnwidth]{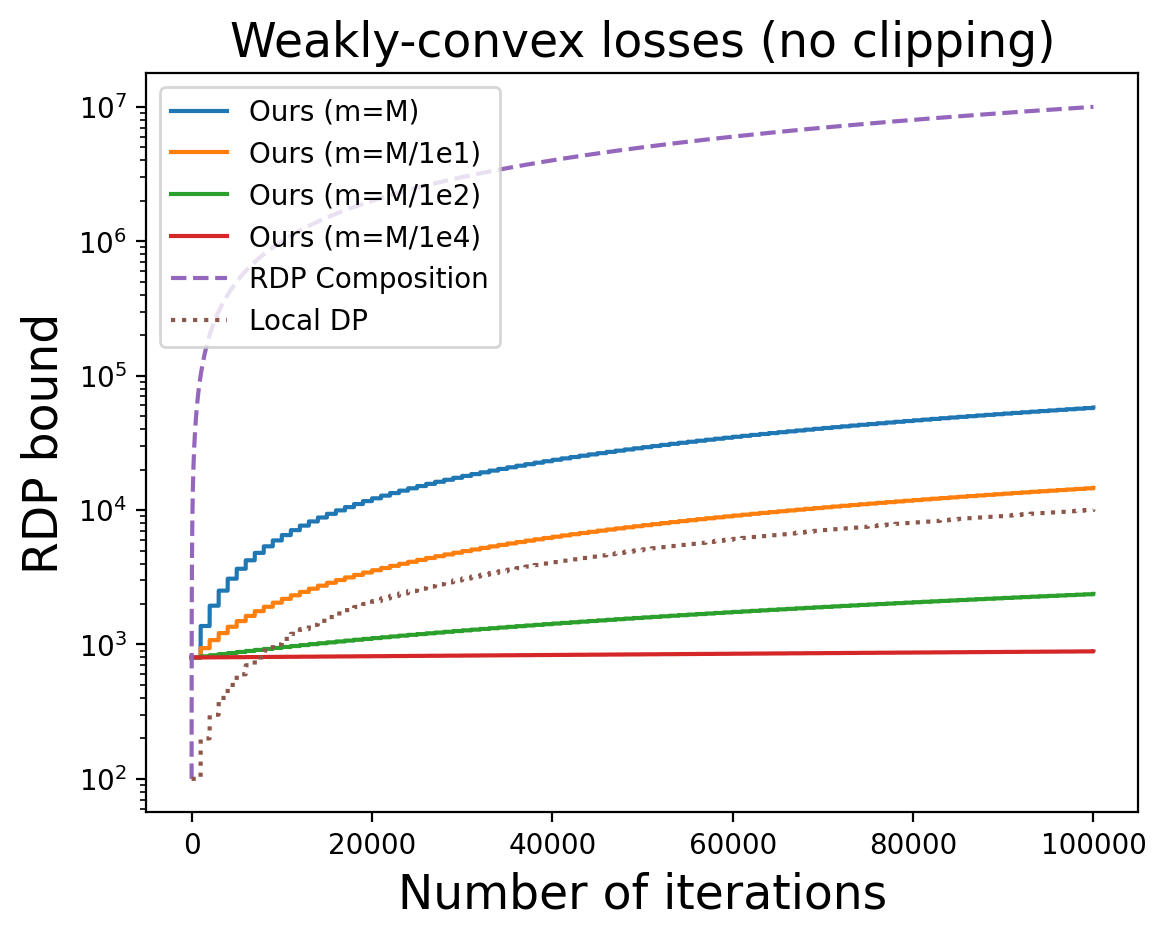}\includegraphics[width=0.4\columnwidth]{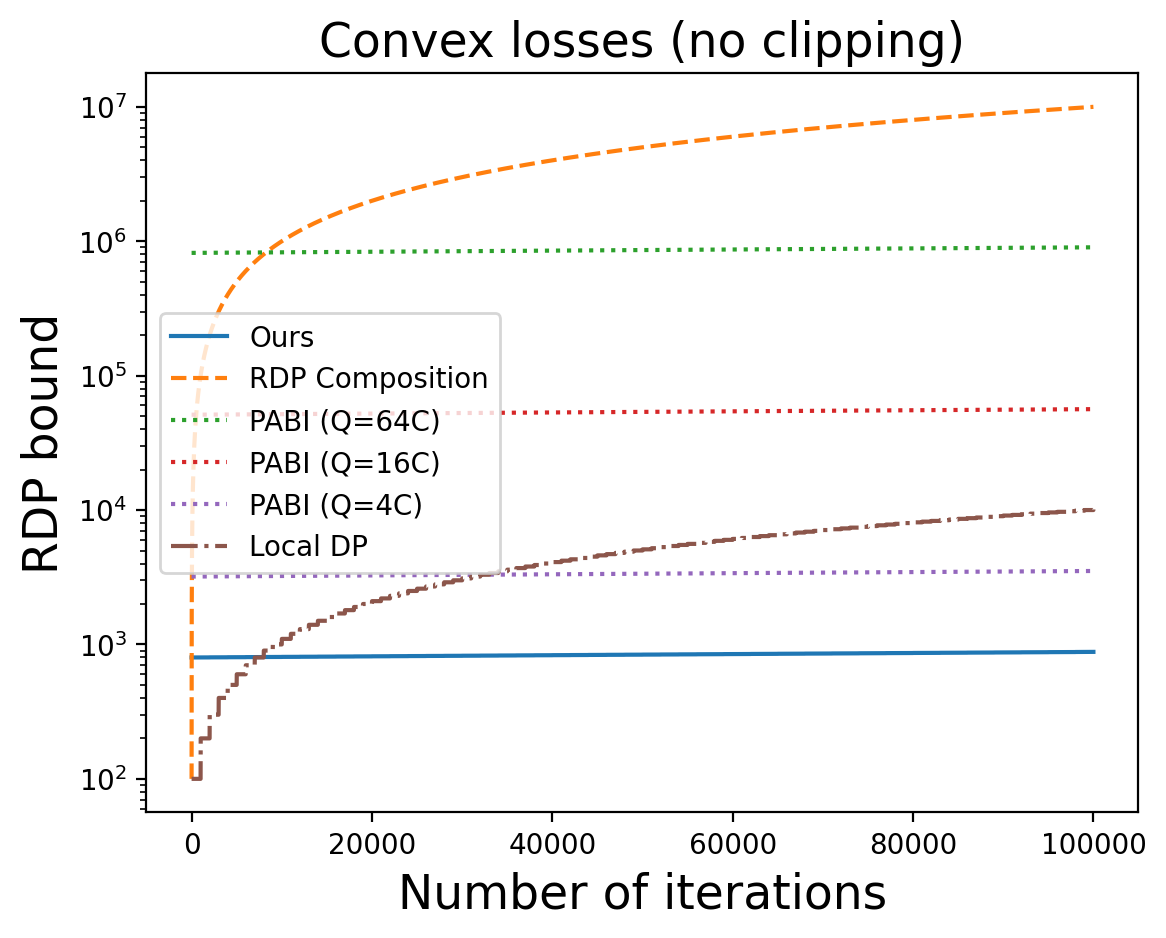}
\hspace*{\fill}
\end{centering}
\caption{Log-scale last-iterate RDP bounds for Algorithm~\ref{alg:dp_sgd} as a function of number
of iterations. The fixed algorithmic parameters
are $\lambda=10^{-5}$, $C=10$, $\sigma=10^{-5}$, $T=10^{5}$, $b=10^{1}$,
and $k=10^{4}$. The free parameters $m$, $M$, and $Q$ are
the weak convexity, Lipschitz-smoothness, and Lipschitz constants
of $f$, respectively, while $C$ is the clipping norm. The\emph{
left }plot considers weakly-convex losses under the
settings of no gradient clipping.
The \emph{right} plot considers the convex case where $Q$ may differ
from $C$. The RDP composition bounds are from  \cite{kaissis2021unified}, the PABI bounds are from \cite{feldman2018privacy}, and the local DP bounds are from \cite{CGK24}.}
\label{fig:bd_graph_summary}
\end{figure}

\vspace*{1em}
\noindent\textbf{Summary tables and graphs}. Table~\ref{tab:rdp_assumptions} lists and labels some assumptions that are commonly used in the RDP literature.
 Table~\ref{tab:bd_summary} compares our bounds against other RDP bounds.
Note that the multi-epoch noisy-SGD algorithm in \cite[Algorithm~1]{feldman2018privacy} only considers the case where the number of dataset passes equals the number of batches in dataset pass and does not consider batched gradients. As a consequence of the latter, its corresponding RDP upper bound does not depend on the batch size $b$.  Figure~\ref{fig:bd_graph_summary} compares our bounds against other bounds as function of the number of iterations performed, under various settings. Note that we do not consider the multi-epoch bound in \cite[Theorem A.1]{CL24} as it requires solving $T$ nonlinear programming problems, and we do not compare the with the bound in \cite{mironov2019r} because it exploits the fact that batches are randomly sampled (whereas our bounds assume the input batches must be obtained deterministically).

\begin{table}[!ht]
\begin{centering}
\begin{tabular}{cl}
\toprule 
\textbf{\footnotesize{}Label} & \textbf{\footnotesize{}Description}\tabularnewline
\midrule
{\footnotesize{}${\cal I}_{c}$} & {\footnotesize{} The regularizer $h$ is the indicator of a closed convex set}\tabularnewline
{\footnotesize{}${\cal H}$} & {\footnotesize{}The domain of a regularizer $h$, $\dom h$, is bounded with diameter $d_{h}$}\tabularnewline
\midrule
{\footnotesize{}${\cal B}_{1}$} & {\footnotesize{}input data batches are of size one }\tabularnewline
{\footnotesize{}${\cal D}$} & {\footnotesize{}input data batches are disjoint }\tabularnewline
{\footnotesize{}${\cal P}$} & {\footnotesize{}input data batches are obtained using Poisson subsampling with
sampling rate $1/\ell$}\tabularnewline
\midrule
{\footnotesize{}${\cal S}_{Q}^{0}$} & {\footnotesize{}$f_{i}$ is $Q$-Lipschitz continuous for every $i\in[k]$, and $Q$ is known }\tabularnewline
{\footnotesize{}${\cal S}_{L}^{1}$} & {\footnotesize{}$\nabla f_{i}$ is $L$-Lipschitz continuous for every
$i\in[k]$, and $L$ is known}\tabularnewline
{\footnotesize{}${\cal R}_{H,\eta}^{1}$} & {\footnotesize{}$\nabla f_{i}$ is $(H,\eta)$-H\"older continuous for every
$i\in[k]$, and $H$ and $\eta$ are known}\tabularnewline
{\footnotesize{}$\Lambda$} & {\footnotesize{}the stepsize needs to be small relative to certain
smoothness constants}\tabularnewline
{\footnotesize{}${\cal A}$} & {\footnotesize{}the given RDP bounds only hold for a small range of values of $\alpha$
and $\sigma$}   \tabularnewline
\midrule
{\footnotesize{}${\cal N}$} & {\footnotesize{}no gradient clipping is applied or the global sensitivity is known}\tabularnewline
{\footnotesize{}$\tilde{{\cal N}}$} & {\footnotesize{}when $\phi$ is convex, no gradient clipping is applied or the global sensitivity is known}\tabularnewline
\bottomrule
\end{tabular}
\par\end{centering}
\caption{List of common assumptions used in RDP accounting literature. For
conciseness we let $\ell$ denote the number of iterations/batches
in a dataset pass and $\protect\dom h$ denote the domain of $h$.\label{tab:rdp_assumptions}}
\end{table}

\begin{table}[!ht]
\begin{centering}
\renewcommand*{\arraystretch}{1.5}
\begin{tabular}{c|>{\centering}p{3cm}c|>{\centering}p{2.8cm}}
\toprule 
\multirow{2}{*}{{\footnotesize{}\textbf{Source}}} & \multicolumn{2}{c|}{{\footnotesize{}\textbf{Asymptotic $(\alpha, \epsilon)-$RDP upper bounds}}} & {\footnotesize{}\textbf{Assumptions}}\tabularnewline
\cline{2-3} \cline{3-3} 
 & \scriptsize{}Convex $\phi$ & \scriptsize{}Nonconvex $\phi$ & {\scriptsize{}(see Table~\ref{tab:rdp_assumptions})}
\tabularnewline
\midrule
{\scriptsize{}\cite[Theorem 3.1]{CGK24}\footnote{ The bound follows directly from Theorem 8 in \cite{BW18} }} & {\scriptsize{}$\dfrac{\alpha E}{2}\left[ \dfrac{\lambda C}{\sigma b}\right]^2$ } & {\scriptsize{}same as convex} & {\scriptsize{}${\cal D}$}\tabularnewline
{\scriptsize{}\cite[Section 3.3]{mironov2019r}} & {\scriptsize{}numerical procedure\tablefootnote{Obtained by evaluating an integral using numerical quadrature techniques.}} & {\scriptsize{}same as convex} & {\scriptsize{}${\color{red}{\cal P}}$}\tabularnewline
{\scriptsize{}\cite[Theorem~11]{mironov2019r}} & {\scriptsize{}$\dfrac{\alpha E}{\ell}\left[\dfrac{\lam C}{b\sigma}\right]^{2}$} & {\scriptsize{}same as convex} & {\scriptsize{}${\color{red}{\cal A}},{\color{red}{\cal P}}$}\tabularnewline
{\scriptsize{}\cite[Theorem~35]{feldman2018privacy}} & {\scriptsize{}$\dfrac{\alpha E}{\ell}\left[\dfrac{\lam Q}{\sigma}\right]^{2}$} & {\scriptsize{}none} & {\scriptsize{}${\cal I}_{c},{\cal N},\Lambda,S_{Q}^{0},S_{L}^{1},{\color{red}{\cal B}_{1}}$}\tabularnewline
{\scriptsize{}\cite[Theorem 3.1]{altschuler2022privacy}} & {\scriptsize{}$\alpha\left[\dfrac{\lam Q}{\sigma}\right]^{2}\min\left\{ \dfrac{E}{\ell},\dfrac{d_{h}}{Q\lam k}\right\} $} & {\scriptsize{}none} & {\scriptsize{}${\cal I}_{c},{\cal N},\Lambda,S_{Q}^{0},S_{L}^{1},{\cal H},{\color{red}{\cal A}}$}\tabularnewline
{\scriptsize{}\cite[Theorem~A.1]{CL24}} & {\scriptsize{}numerical procedure\tablefootnote{\label{fn:1}Obtained by solving $T$ nonlinear programming problems.}} & {\scriptsize{}numerical procedure\tablefootnote{Same procedure as in the nonconvex case, but with different parameters.}} & {\scriptsize{}${\cal I}_{c},\tilde{{\cal N}},\Lambda,{S}_{Q}^{0},{\cal R}_{H,\eta}^{1},{\cal H}$}\tabularnewline
\midrule
\multirow{3}{*} {\textbf{\scriptsize{}Ours}} & {\scriptsize{}$\dfrac{\alpha}{\sigma^{2}}\left[L_{\lam}d_{h}+\dfrac{\lam C}{b}\right]^{2}$} & {\scriptsize{}same as convex} & {\scriptsize{}$\Lambda,S_{L}^{1},{\cal H}, {\cal D}$}\tabularnewline
 & {\scriptsize{}$\dfrac{\alpha E}{\ell}\left[\dfrac{\lam C}{b\sigma}\right]^{2}$} & {\footnotesize{}$\alpha E\theta_{L_{\lam}}(\ell)\left[\dfrac{\lam C}{\sigma}\right]^{2}$} & {\scriptsize{}$\Lambda,S_{L}^{1},{\cal N}, {\cal D}$}\tabularnewline
 & {\scriptsize{}$\alpha E\theta_{\sqrt{2}}(\ell)\left[\dfrac{\lam C}{\sigma}\right]^{2}$} & {\footnotesize{}$\alpha E\theta_{\sqrt{2}L_{\lam}}(\ell)\left[\dfrac{\lam C}{\sigma}\right]^{2}$} & {\scriptsize{}$\Lambda,S_{L}^{1}, , {\cal D}$}\tabularnewline
\bottomrule
\end{tabular}
\par\end{centering}
\caption{Asymptotic $\varepsilon$ upper bounds for $(\alpha, \varepsilon)$-R\'enyi differential privacy of the
last iterate after $T$ iterations of DP-SGD.
Here, $\lam$ is the stepsize, $C$ is the clipping norm, $\sigma$ is the standard deviation
of the Gaussian noise, $\ell$ is the number of iterations in one
dataset pass, and $E$ is the number of dataset passes. Also, $L_{\lam}:=\sqrt{1+\lam \kappa m}$
for some $\kappa \leq 4$ and weak-convexity parameter $m$ (see (\ref{eq:lipschitz})), and
$\theta_{L}(\ell):=L^{2(\ell-1)}/\sum_{i=0}^{\ell-1}L^{2i}$. Particularly strong assumptions are highlighted
in {\color{red} red}.}
\label{tab:bd_summary}
\end{table}

% \mr{${\cal B}_1 \to$ To avoid overriding the ball, $\cal {Y}_1$? [wild: 1 becomes $\infty$ haha]}

% \mr{When I do the conversion of theorem 11 in \cite{mironov2019r} I get $2\alpha T\left[\frac{b}{\sigma}\right]$. We have that $q^2$ is $b/k$, $\sigma = \frac{C}{n}$. And then we compose $T$ times. Converting to $(\epsilon, \delta)-$DP we get it scales as $\sqrt{T}$.}

% \mr{$\cal A$ seems to be a different category than $\cal B$ and $\cal P$.  }

\vspace*{1em}
\noindent\textbf{Technical nuances}.
The bound in \eqref{eq:intro_bd} with ${\cal B}_\lam$ as in \eqref{eq:smooth_b} and $L_\lam=1$ might appear to follow from subsampled RDP composition results such as \cite{mironov2019r}. 
However, those results only apply to DP-SGD variants where the batches \textit{are sampled randomly}, an assumption that does not hold when  batches are cyclically traversed. While established Python libraries like Opacus \cite{opacus} and TensorFlow Privacy \cite{tensorflow-privacy} implement and account for random sampled batches (such as those obtained by Poisson subsampling), 
these implementations address a different issue.  One has to ensure the optimizer truly samples at random from a pre-specified distribution, which becomes incredibly difficult with large-scale datasets (see Appendix~\ref{app:random_batch_sampling} for an extended discussion).  Consequently, any privacy guarantee predicated on this idealized random sampling assumption becomes effectively meaningless when the actual optimization process deviates from it (as is the case with cyclical batch traversal). It is worth mentioning that DP-FTRL \cite{KMSTTX21} was specifically developed to address this gap and the method accepts a potentially weaker DP guarantee in exchange for practical applicability.

For the special case where (i) only one dataset pass is performed, (ii) the objective function is nonconvex , and (iii) each $\nabla f_i$ is Lipschitz continuous, the bound in \eqref{eq:intro_bd} with ${\cal B}_\lam$ as in \eqref{eq:smooth_b} holds with $T = \ell$. Consequently, we obtain an RDP bound that does not grow with the number of iterations --- in contrast to the RDP composition bounds in \cite{kaissis2021unified} which do scale linearly in $T$ or $E$, depending on the sampling assumption. Note that \eqref{eq:dp_sgd_bd1} and Theorem~\ref{thm:general_compl} show that as $\theta_{L_\lam}(\cdot)\downarrow1$, our bounds converge to an expression that depends linearly on $E/\ell$, matching the bound for PABI. Figure~\ref{fig:bd_graph_summary} illustrates the regimes under which we improve previous work.

The addition of the composite term $h(\cdot)$ in \eqref{eq:main_prb} is not a trivial extension and greatly complicates the analysis. For example, the objective function $\phi$ in \eqref{eq:main_prb} is no longer differentiable in general (even on $\dom \phi$), and more general analyses must be used to handle this nonsmoothness. 
Under stepsize $\lam$ and $L$-Lipschitz continuous $\nabla f_i$ for $i\in [k]$, paper \cite{CL24} shows that the DP-SGD update in the nonconvex case is $(1+\lam L)$-Lipschitz continuous, which is independent of the weak convexity constant. Consequently, the established RDP bounds in \cite{CL24} in the setting where the weak convexity constant $m$ is positive but near zero (i.e., the function is nearly convex) may be an overestimate of the true RDP bound.

For the convex case, it is worth emphasizing that we do not require each to be Lipschitz continuous case in order to  bound $\nabla f_i$ (see, for example, \cite{feldman2018privacy, altschuler2022privacy, feldman2020private, CL24} which do require this assumption). 
As a consequence, our analysis is applicable to a substantially wider class of objective functions. 
Moreover, all other existing bounds in the literature of the form in \eqref{eq:intro_bd} replace the parameter $C$ with a Lipschitz constant $Q$ of $\phi(x)$ from \eqref{eq:main_prb}, and these bounds are generally proportional to $Q^2$. Consequently, when $C \ll Q$, e.g., when $\phi(x)$ is a quadratic function on a large compact domain, our bounds are significantly tighter (see Figure~\ref{fig:bd_graph_summary} for an illustration).

\vspace*{1em}
\noindent\textbf{Organization}. The remainder of the paper gives a formal presentation of the results, including the key assumptions on \eqref{eq:main_prb}, the topological properties of the DP-SGD update operator, and the non-asymptotic RDP bounds on the last DP-SGD iterates. 

\section{Privacy bounds for DP-SGD}

\label{sec:privacy_bds}

This section formally presents the main RDP bounds for the last iterates of Algorithm~\ref{alg:dp_sgd}. For conciseness, the lengthy proofs of the main results are given in Appendix~\ref{app:main_results}.

We start by precisely giving the assumptions on \eqref{eq:main_prb}. Given $h:\rn\mapsto(-\infty,\infty]$ and $f_{i}:\dom h\mapsto\r$
for $i\in[k]$, consider the following assumptions:
\vspace*{1em}
\begin{itemize}
\item[\color{violet}(A1)] $h\in\cConv(\rn)$;
\item[\color{violet}(A2)] there exists $m,M\geq 0$ such that for $i\in[k]$ the function
$f_{i}$ is differentiable on an open set containing $\dom h$ and 
\begin{equation}
-\frac{m}{2}\|x-y\|^2\leq f_{i}(x)-f_{i}(y)-\left\langle \nabla f_{i}(y),x-y\right\rangle \leq\frac{M}{2}\|x-y\|^{2} \quad \forall x,y\in \dom h.\label{eq:lipschitz}
\end{equation}
\end{itemize}
We now give a few remarks about (A1)--(A2). First, (A1) is necessary to ensure that $\prox_h(\cdot)$ is well-defined. Second, it can be shown\footnote{See, for example, \cite{beck2017first,nesterov2018lectures} and \cite[Proposition 2.1.55]{kong2021accelerated}.} that (A2) is equivalent to the assumption that $\nabla f$ is $M$-Lipschitz continuous when $m=M$. Third, the lower bound in \eqref{eq:lipschitz} is equivalent to assuming that $f_i(\cdot)+m\|\cdot\|^2/2$ is convex and, hence, if $m=0$ then $f_i$ is convex. The parameter $m$ is often called a weak-convexity parameter of $f_i$. Fourth, using symmetry arguments and the third remark, if $M=0$ then $f_i$ is concave. 
Finally, the third remark motivates why we choose two parameters, $m$ and $M$, in \eqref{eq:lipschitz}. Specifically, we use $m$ (resp. $M$) to develop results that can be described in terms of the level of convexity (resp. concavity) of the problem. 

We now develop the some properties of an SGD-like update. Given $\{q_{i}\}\subseteq\cConv(\rn)$ with $\dom q_{i}\subseteq\dom h$ and $B\subseteq[k]$, define the prox-linear operator
\begin{equation}
{\cal A}_{\lam}(x) = {\cal A}_{\lam}(x,\{f_{i}\},\{q_{i}\}):=x - \frac{\lam}{|B|}\sum_{i\in B}\prox_{q_{i}}(\nabla f_{i}(x)).\label{eq:gen_dp_sgd_update}
\end{equation}
Clearly, when $\prox_{q_i}(y)=y$ for every $y\in\rn$, the above update corresponds to a SGD step applied to the problem of minimizing $\sum_{i=1}^k f_i(z)$ (with respect to $z$) under the stepsize $\lam$ and starting point $x$. Moreover, while it is straightforward to show that ${\cal A}_{\lam}(\cdot)$ is $(1+\lam \max\{m,M\})$-Lipschitz continuous when $\{f_i\}$ satisfies (A2)\footnote{See, for example, \cite[Appendix A.6]{CL24}.}, we prove in Proposition~\ref{prop:prox_lipsch}(b) that the Lipschitz constant can be improved to $\sqrt{1+ \kappa\lam m}$ for some $\kappa \leq 4$ when $\lam$ is small. Notice that the former constant does not converge to one when $m\to0$, e.g., when $f_i$ becomes more convex, while the latter one does.

We are now present some important properties of ${\cal A}_\lam(\cdot)$.
\begin{proposition}
\label{prop:prox_lipsch}
Let $(m,M)$ be as in assumption (A2), and define
\begin{equation}
L_{\lam} = L_\lam(m,M) :=\sqrt{1 + 2\lam m\left[1+\frac{m}{2(M+m)}\right]}\quad\forall\lam>0.
\label{eq:L_lam}
\end{equation}
Then, the following statements hold:
\begin{itemize}
\item[(a)]if $\dom q_i$ is bounded with diameter $C$ for $i\in[k]$, then for any $\lam > 0$ we have
\begin{equation}
    \|{\cal A}_{\lam}(x)-{\cal A}_{\lam}(y)\|\leq \|x-y\|+2\lam C\quad\forall x,y\in\dom h;
\end{equation}
\item[(b)]if $\{f_i\}$ satisfies (A2) and $\lam\leq 1/[2(M+m)]$ then ${\cal A}_{\lam}(\cdot)$ is $\sqrt{2} L_\lam$-Lipschitz continuous;
\item[(c)]if $\{f_i\}$ satisfies (A2),  $\lam\leq 1/(M+m)$, and 
$\nabla f_i(x) = \prox_{q_i}(\nabla f_i(x))$ for every $i \in [k]$ and $x\in\dom h$, 
then ${\cal A}_{\lam}(\cdot)$ is $L_\lam$-Lipschitz continuous on $\dom h$.
\end{itemize}
\end{proposition}

Some remarks are in order. First, $L_\lam(0,M) = 1$ and, hence, ${\cal{A}}_\lam(\cdot)$ is nonexpansive when $f_i$ is convex for every $i\in[k]$, $\lam \leq 1/M$, and $\nabla f(x_i) = \prox_{q_i}(\nabla f(x_i))$. 
Second, if $\lam=1/(2m)$ then $L_\lam(m,0)=\sqrt{3}$ and, hence, ${\cal{A}}_\lam(\cdot)$  $\sqrt{6}$-Lispchitz continuous when $f_i$ is concave.
Third, like the first remark, $L_0(m,M) = 1$ implies that ${\cal{A}}_\lam(\cdot)$ is nonexpansive. Finally, when $m=M$ and $\lam=1/(2m)$, we have $L_\lam(m,M)=\sqrt{5/2}$ and ${\cal{A}}_\lam(\cdot)$ is $\sqrt{5}$-Lispchitz continuous.

For the remainder of this section, suppose $h$ satisfies (A1) and let $f_{i}':\rn\mapsto(-\infty,\infty]$
for $i\in[k]$ be such that there exists $i^{*}\in[k]$ where $f_{i}'=f_{i}$ for every $i\neq i^{*}$
and $f_{i^{*}}'\neq f_{i^{*}}$, i.e., $\{f_i\} \sim \{f_i'\}$. That is, $i^*$ is the index where the neighboring datasets $\{f_i\}$ and $\{f_i'\}$ differ. 

We also make use of the follow assumption.
\vspace*{1em}
\begin{itemize}
\item[\color{violet}(A3)] The functions $\{f_{i}'\}$ satisfy assumption (A2)
with $f_{i}=f_{i}'$ for every $i\in[k]$.
\end{itemize}
\vspace*{1em}

We now present the main RDP bounds in terms of the following constants:
\begin{equation}
\label{eq:dh_theta_def}
d_{h}:=\sup\{\|x-y\|:x,y\in\dom h\}, 
\quad 
\theta_{L}(0):=0,\quad
\theta_{L}(s):=\frac{L^{2(s-1)}}{\sum_{j=0}^{s-1}L^{2j}} \quad \forall s\geq 1.
\end{equation} 
We first present a bound where $\dom h$ is bounded with diameter $d_h < \infty$.
\begin{theorem}
\label{thm:dp_sgd_bd1}
Let $\{X_{t}\}$ and $\{X_{t}'\}$
denote the iterates generated by Algorithm\ \ref{alg:dp_sgd} with per-example loss
functions $\{f_{i}\}$ and $\{f_{i}'\}$, respectively, and fixed $\lam$, $C$, $b$, $\sigma$, $\{N_{t}\}$, $T$, and $X_{0}$ for both sequences of iterates. If $\lam\leq1/[2(m+M)]$ and (A1)--(A3) hold, then
\begin{align}
D_{\alpha}(X_{T}\|X_{T}') & \leq\frac{\alpha}{2\sigma^{2}}\left(L_\lam d_{h}+\frac{2\lam C}{b}\right)^{2},\label{eq:dp_sgd_bd1}
\end{align} 
where $L_{\lam}$ and $d_h$ are as in \eqref{eq:L_lam} and \eqref{eq:dh_theta_def}, respectively.
\end{theorem}

We now present the RDP bounds for when $\dom h$ is (possibly) unbounded.
\begin{theorem}
\label{thm:general_compl}
Let $\{X_{t}\}$, $\{X_{t}'\}$, $\lam$, $\sigma$, $b$,
$C$, and $T$ be as in Theorem~\ref{thm:dp_sgd_bd1}, and denote $\ell:=k/b$ and $E:=\lfloor T/\ell\rfloor$.
If $\lam\leq1/[2(m+M)]$ and (A1)--(A3) hold, then 
\begin{equation}
D_{\alpha}(X_{T}\|X_{T}')\leq 4\alpha\left(\frac{\lam C}{b\sigma}\right)^{2}\left[1 + E \cdot \theta_{\sqrt{2} L_\lam}(\ell)\right],\label{eq:dp_sgd_indep_bd1}
\end{equation}
where $L_{\lam}$ and $\theta_L(\cdot)$ are as in \eqref{eq:L_lam} and \eqref{eq:dh_theta_def}, respectively. On the other hand, if 
\begin{equation}
\max_{i\in[k], t\in[T]}\{\|\nabla f_i(X_t)\|, \|\nabla f_i(X_t')\|\} \leq C, 
\label{eq:noclip_cond}
\end{equation}
i.e., no gradient clipping is performed, and $\lam \leq {1}/{(M+m)}$ then  
\begin{equation}
D_{\alpha}(X_{T}\|X_{T}')\leq 4\alpha\left(\frac{\lam C}{b\sigma}\right)^{2}\left[1 + E \cdot \theta_{L_\lam}(\ell)\right] .\label{eq:dp_sgd_indep_bd2}
\end{equation}
\end{theorem}

We conclude with a few remarks about the above bounds. First, the bound in \eqref{eq:dp_sgd_indep_bd2} is on the same order of magnitude as the bound in \cite{feldman2018privacy} in terms of $T$ and $\ell$ when $L_\lam=1$. However, the right-hand-side of \eqref{eq:dp_sgd_indep_bd2} scales linearly with a $\lam^2$ term, which does not appear in \cite{feldman2018privacy}. Second, as $\theta_{L_\lam}(\cdot)\leq 1$, the right-hand-sides of \eqref{eq:dp_sgd_indep_bd1} and \eqref{eq:dp_sgd_indep_bd2} increases (at most) linearly with respect to the number of dataset passes $E$. 
Third, substituting $\sigma = \Theta({C/[b\sqrt{\epsilon}]})$ in \eqref{eq:dp_sgd_indep_bd1} yields a bound that depends linearly on $\varepsilon$ and is invariant to changes in $C$ and $b$.
In Appendix~\ref{app:params}, we discuss further choices of the parameters in \eqref{eq:dp_sgd_indep_bd2} and their properties.

\bibliographystyle{plain}
\bibliography{privacy_amplification}

\begin{thebibliography}{10}

\bibitem{ACGM+16}
Martin Abadi, Andy Chu, Ian Goodfellow, H.~Brendan McMahan, Ilya Mironov, Kunal
  Talwar, and Li~Zhang.
\newblock Deep learning with differential privacy.
\newblock In {\em ACM SIGSAC Conference on Computer and Communications
  Security}, 2016.

\bibitem{altschuler2022privacy}
Jason Altschuler and Kunal Talwar.
\newblock {P}rivacy of noisy stochastic gradient descent: {M}ore iterations
  without more privacy loss.
\newblock {\em Advances in Neural Information Processing Systems (NeurIPS)},
  2022.

\bibitem{ABT24}
Jason~M. Altschuler, Jinho Bok, and Kunal Talwar.
\newblock On the privacy of noisy stochastic gradient descent for convex
  optimization.
\newblock {\em SIAM Journal on Computing}, 2024.

\bibitem{asoodeh2023privacy}
Shahab Asoodeh and Mario D\'iaz.
\newblock Privacy loss of noisy stochastic gradient descent might converge even
  for non-convex losses.
\newblock {\em arXiv preprint arXiv:2305.09903}, 2023.

\bibitem{BW18}
Borja Balle and Yu-Xiang Wang.
\newblock Improving the {Gaussian} mechanism for differential privacy:
  Analytical calibration and optimal denoising.
\newblock In {\em International Conference on Machine Learning (ICML)}. PMLR,
  2018.

\bibitem{BST14}
Raef Bassily, Adam Smith, and Abhradeep Thakurta.
\newblock Private empirical risk minimization: Efficient algorithms and tight
  error bounds.
\newblock In {\em Symposium on foundations of computer science}, 2014.

\bibitem{beck2017first}
Amir Beck.
\newblock {\em First-order methods in optimization}.
\newblock SIAM, 2017.

\bibitem{BSA24}
Jinho Bok, Weijie Su, and Jason~M Altschuler.
\newblock Shifted interpolation for differential privacy.
\newblock {\em arXiv preprint arXiv:2403.00278}, 2024.

\bibitem{CD24}
Xufeng Cai and Jelena Diakonikolas.
\newblock Last iterate convergence of incremental methods and applications in
  continual learning.
\newblock {\em arXiv preprint arXiv:2403.06873}, 2024.

\bibitem{CMS11}
Kamalika Chaudhuri, Claire Monteleoni, and Anand~D Sarwate.
\newblock Differentially private empirical risk minimization.
\newblock {\em Journal of Machine Learning Research}, 2011.

\bibitem{CL24}
Eli Chien and Pan Li.
\newblock Convergent privacy loss of noisy-{SGD} without convexity and
  smoothness.
\newblock {\em arXiv preprint arXiv:2410.01068}, 2024.

\bibitem{CYS21}
Rishav Chourasia, Jiayuan Ye, and Reza Shokri.
\newblock Differential privacy dynamics of {L}angevin diffusion and noisy
  gradient descent.
\newblock In {\em Advances in Neural Information Processing Systems (NeurIPS)},
  2021.

\bibitem{CGKK+24}
Lynn Chua, Badih Ghazi, Pritish Kamath, Ravi Kumar, Pasin Manurangsi, Amer
  Sinha, and Chiyuan Zhang.
\newblock How private are {DP}-{SGD} implementations?
\newblock In {\em International Conference on Machine Learning (ICML)}, 2024.

\bibitem{CGK24}
Lynn Chua, Badih Ghazi, Pritish Kamath, Ravi Kumar, Pasin Manurangsi, Amer
  Sinha, and Chiyuan Zhang.
\newblock Scalable {DP-SGD}: Shuffling vs. poisson subsampling.
\newblock In {\em Conference on Neural Information Processing Systems
  (NeurIPS)}, 2024.

\bibitem{DR14}
Cynthia Dwork, Aaron Roth, et~al.
\newblock The algorithmic foundations of differential privacy.
\newblock {\em Foundations and Trends{\textregistered} in Theoretical Computer
  Science}, 2014.

\bibitem{feldman2020private}
Vitaly Feldman, Tomer Koren, and Kunal Talwar.
\newblock Private stochastic convex optimization: optimal rates in linear time.
\newblock In {\em ACM SIGACT Symposium on Theory of Computing (STOC)}, 2020.

\bibitem{feldman2018privacy}
Vitaly Feldman, Ilya Mironov, Kunal Talwar, and Abhradeep Thakurta.
\newblock Privacy amplification by iteration.
\newblock In {\em Annual Symposium on Foundations of Computer Science (FOCS)},
  2018.

\bibitem{KMSTTX21}
Peter Kairouz, Brendan McMahan, Shuang Song, Om~Thakkar, Abhradeep Thakurta,
  and Zheng Xu.
\newblock Practical and private (deep) learning without sampling or shuffling.
\newblock In {\em International Conference on Machine Learning}, pages
  5213--5225. PMLR, 2021.

\bibitem{kairouz2015composition}
Peter Kairouz, Sewoong Oh, and Pramod Viswanath.
\newblock The composition theorem for differential privacy.
\newblock In {\em International {C}onference on {M}achine {L}earning (ICML)},
  2015.

\bibitem{kaissis2021unified}
Georgios Kaissis, Moritz Knolle, Friederike Jungmann, Alexander Ziller, Dmitrii
  Usynin, and Daniel Rueckert.
\newblock A unified interpretation of the gaussian mechanism for differential
  privacy through the sensitivity index.
\newblock {\em arXiv preprint arXiv:2109.10528}, 2021.

\bibitem{kong2021accelerated}
Weiwei Kong.
\newblock Accelerated inexact first-order methods for solving nonconvex
  composite optimization problems.
\newblock {\em arXiv preprint arXiv:2104.09685}, 2021.

\bibitem{liang2024unified}
Jiaming Liang and Renato~D.C. Monteiro.
\newblock A unified analysis of a class of proximal bundle methods for solving
  hybrid convex composite optimization problems.
\newblock {\em Mathematics of Operations Research}, 49(2):832--855, 2024.

\bibitem{LZ24}
Zijian Liu and Zhengyuan Zhou.
\newblock On the last-iterate convergence of shuffling gradient methods.
\newblock In {\em International Conference on Machine Learning (ICML)}, 2024.

\bibitem{tensorflow-privacy}
Google LLC.
\newblock Tensorflow privacy.
\newblock \url{https://github.com/tensorflow/privacy}, 2019.

\bibitem{mironov2019r}
Ilya Mironov, Kunal Talwar, and Li~Zhang.
\newblock R\'enyi differential privacy of the sampled {G}aussian mechanism.
\newblock {\em arXiv preprint arXiv:1908.10530}, 2019.

\bibitem{nesterov2018lectures}
Yurii Nesterov et~al.
\newblock {\em Lectures on convex optimization}, volume 137.
\newblock Springer, 2018.

\bibitem{SCS13}
Shuang Song, Kamalika Chaudhuri, and Anand~D Sarwate.
\newblock Stochastic gradient descent with differentially private updates.
\newblock In {\em IEEE Global Conference on Signal and Information Processing}.
  IEEE, 2013.

\bibitem{YS22}
Jiayuan Ye and Reza Shokri.
\newblock Differentially private learning needs hidden state (or much faster
  convergence).
\newblock {\em Advances in Neural Information Processing Systems (NeurIPS)},
  2022.

\bibitem{opacus}
Ashkan Yousefpour, Igor Shilov, Alexandre Sablayrolles, Davide Testuggine,
  Karthik Prasad, Mani Malek, John Nguyen, Sayan Ghosh, Akash Bharadwaj,
  Jessica Zhao, Graham Cormode, and Ilya Mironov.
\newblock Opacus: {U}ser-friendly differential privacy library in {PyTorch}.
\newblock {\em arXiv preprint arXiv:2109.12298}, 2021.

\end{thebibliography}

\newpage
\appendix

\section{Derivation of main results}

\label{app:main_results}

This appendix derives the main results, namely, Theorems~\ref{thm:dp_sgd_bd1} and \ref{thm:general_compl}. It contains three subappendices. The first one derives important properties of a family of randomized operators, the second specializes these results to the DP-SGD update operator in \eqref{eq:gen_dp_sgd_update}, and the last one gives the proofs of Theorems~\ref{thm:dp_sgd_bd1} and \ref{thm:general_compl} using the previous two subappendices.

\subsection{General operator analysis}
\label{sec:rand_op}

This subappendix gives some crucial properties about randomized proximal Lipschitz operators, which consist of evaluating a Lipschitz proximal operator followed by adding Gaussian noise. More specifically, it establishes several RDP bounds based on the closeness of neighboring operators.

We first bound the shifted R\'enyi divergence of a randomized proximal Lipschitz operator. The proof of this result is a straightforward extension of the argument in \cite[Theorem 22]{feldman2018privacy} 
from $1$-Lipschitz operators to $L$-Lipschitz operators with additive residuals.

To begin, we present two calculus rules for the shifted R\'enyi divergence
given in \eqref{eq:shifted_div}. In particular, the proof of the
second rule is a minor modification of the proof given for \cite[Lemma 21]{feldman2018privacy}.
\begin{lemma}
\label{lem:shifted_calc}
For random variables $\{X',X,Z\}$ and $a,s\geq0$ and
$\alpha\in(1,\infty)$, it holds that
\begin{itemize}
\item[(a)] $D_{\alpha}^{(\tau)}(X+Z\|X'+Z)\leq D_{\alpha}^{(\tau+a)}(X\|X')+\sup_{c\in\rn}\{D_{\alpha}([Z+c]\|Z):\|c\|\leq a\}$;
\item[(b)] for some $L,\zeta>0$, if $\phi'$ and
$\phi$ satisfy 
\[
\sup_{u}\|\phi'(u)-\phi(u)\|\leq s,\quad \|\Phi(x) - \Phi(y)\|\leq L\|x-y\| + \zeta, \quad \forall \Phi\in\{\phi,\phi'\},
\]
for any $x,y\in\dom\phi\cap\dom\phi'$, then
\[
D_{\alpha}^{(L\tau+\zeta+s)}(\phi(X)\|\phi'(X'))\leq D_{\alpha}^{(\tau)}(X\|X').
\]
\end{itemize}
\end{lemma}
\begin{proof}
(a) See \cite[Lemma 20]{feldman2018privacy}.

(b) By the definitions of of $D_{\alpha}^{(\tau)}(\mu\|\nu)$ and
${\cal W}_{\infty}(\mu,\nu)$, there exist a joint distribution $(X,Y)$
such that $D_{\alpha}(Y\|X')=D_{\alpha}^{(\tau)}(X\|X')$ and $\|X-Y\|\leq\tau$
almost surely. Now, the post-processing property of R\'enyi divergence implies that 
\[
D_{\alpha}(\phi(Y)\|\phi'(X'))\leq D_{\alpha}(\phi(Y)\|X') \leq D_{\alpha}(Y\|X') =D_{\alpha}^{(\tau)}(X\|X').
\]
Using our assumptions on $\phi$ and $\phi'$ and the triangle inequality, we then have
\begin{align*}
\|\phi(X)-\phi'(Y)\| & \leq\|\phi(X)-\phi(Y)\|+\|\phi(Y)-\phi'(Y)\| \\
& \leq L\|X-Y\|+ \zeta +s\leq L\tau + \zeta +s,
\end{align*}
almost surely. Combining the previous two inequalities, yields the
desired bound in view of the definitions of $D_{\alpha}^{(\tau)}(\mu\|\nu)$
and ${\cal W}_{\infty}(\mu,\nu)$.
\end{proof}

The next result is the aforemention shifted RDP bound.

\begin{lemma}
\label{lem:main_recurrence}
For some $L,\zeta\geq0$, suppose $\phi'$ and
$\phi$ satisfy \eqref{eq:phi_reg}
for any $x,y\in\dom\phi\cap\dom\phi'$.
Moreover, let $Z\sim{\cal N}(0,\sigma^{2}I)$ and $\psi\in\cConv(\rn)$.
Then, for any scalars $a,\tau\geq0$ and $\alpha\in(1,\infty)$ satisfying
$L\tau+\zeta+s-a\geq0,$ and random variables $Y$ and $Y'$, it holds that
\begin{equation}
D_{\alpha}^{(L\tau+\zeta+s-a)}(\prox_{\psi}(\phi(Y)+Z)\|\prox_{\psi}(\phi'(Y')+Z))\leq D_{\alpha}^{(\tau)}(Y\|Y')+\frac{\alpha a^{2}}{2\sigma^{2}}.\label{eq:main_recurrence}
\end{equation}
\end{lemma}

\begin{proof}
We first have that 
\begin{equation}
\sup_{\tau\in\rn}\{D_{\alpha}([Z+c]\|Z):\|c\|\leq a\}=\sup_{c\in\rn}\left\{ \frac{\alpha c^{2}}{2\sigma^{2}}:\|c\|\leq a\right\} =\frac{\alpha a^{2}}{2\sigma^{2}},\label{eq:R\'enyi_normal}
\end{equation}
from the well-known properties of the R\'enyi divergence. Using \eqref{eq:R\'enyi_normal},
Lemma~\ref{lem:shifted_calc}(a) with $(X,X')=(\phi(Y),\phi'(Y'))$,
and Lemma~\ref{lem:shifted_calc}(b) with $(\phi,\phi',L,s)\in\{(\phi,\phi',L,s),({\rm prox}_{\psi},{\rm prox}_{\psi},1,0)\}$,
we have
\begin{align*}
 & D_{\alpha}^{(L\tau+s-a)}({\rm prox}_{\psi}(\phi(Y)+Z)\|{\rm prox}_{\psi}(\phi'(Y')+Z))\leq D_{\alpha}^{(L\tau+s-a)}(\phi(Y)+Z\|\phi'(Y')+Z)\\
 & \leq D_{\alpha}^{(L\tau+s)}(\phi(Y)\|\phi'(Y'))+\frac{\alpha a^{2}}{2\sigma^{2}}\leq D_{\alpha}^{(\tau)}(Y\|Y')+\frac{\alpha a^{2}}{2\sigma^{2}}.
\end{align*}
\end{proof}

Note that the second inequality in \eqref{eq:phi_reg} is equivalent to $\Phi$ being $L$-Lipschitz continuous when $\zeta=0$, and that the conditions in \eqref{eq:phi_reg} need to only hold on $\dom \phi \cap \dom \phi'$.

We next apply \eqref{eq:main_recurrence} to a sequence of points generated by the update 
\begin{equation}
Y \xleftarrow{} \prox_\psi(\phi(Y)+Z) \label{eq:gen_prox_update}
\end{equation}
under different assumptions on $\zeta$ and $\tau$ and a single dataset pass. Before proceeding, we first present a technical lemma.

\begin{lemma}
\label{lem:tech_bt}
Given scalars $L>1$ and positive integer $T\geq1$, let 
\begin{equation}
S_{T}:=\sum_{i=0}^{T-1}L^{2i},\quad b_{t}:=\left(\frac{L^{T-t}}{S_{T}}\right)L^{T-1},\quad R_{t}:=L^{t-1}-\sum_{i=1}^{t}b_{i}L^{t-i},\quad t\geq0\label{eq:tech_bt}
\end{equation}
Then, for every $t\in[T]$,
\begin{itemize}
\item[(a)] $R_{t+1}=LR_{t}-b_{t+1}$;
\item[(b)] $R_{t}\geq0$ and $R_{T}=0$;
\item[(c)] $\sum_{t=1}^{T}b_{t}^{2}=\theta_{L}(T)$.
\end{itemize}
\end{lemma}
\begin{proof}
Let $t\in[T]$ be fixed.

(a) This is immediate from the definition of $R_{t}$. 

(b) We have that
\begin{align*}
S_{T}R_{t} & =S_{T}\left(L^{t}-\sum_{i=1}^{t}b_{i}L^{t-i}\right)=\sum_{i=0}^{T-1}L^{2i+t-1}-\sum_{i=1}^{t}L^{2T+t-2i-1}=L^{t-1}\left[\sum_{i=0}^{T-1}L^{2i}-\sum_{i=1}^{t}L^{2(T-i)}\right]\\
 & =L^{t-1}\sum_{i=0}^{T-1-t}L^{2i}\geq0.
\end{align*}
Evaluating the above expression at $t=T$ clearly gives $R_{T}=0$.

(c) The case of $T=0$ is immediate. For the case of $T\geq1$, we use the definitions of $b_t$ and $S_T$ to obtain
\begin{align*}
\sum_{t=1}^{T}b_{t}^{2} & =\frac{L^{2(T-1)}\sum_{i=0}^{T-1}L^{2i}}{\left(\sum_{i=0}^{T-1}L^{2i}\right)^{2}}=\frac{L^{2(T-1)}}{S_{T}}=\theta_L(T). 
\end{align*}
\end{proof}

We now present the shifted RDP properties of the update in \eqref{eq:gen_prox_update}. This particular result generalizes the one in  \cite{feldman2018privacy}, which only considers the case of $L=1$ and $\zeta=0$.

\begin{lemma}
\label{lem:single_pass_div}
Let $L,\zeta\geq 0$, $T\geq1$, and $\ell\in[T]$ be fixed. Given $\psi\in\cConv(\rn)$,
suppose $\{\phi_{t}\}_{t=1}^T$, $\{\phi_{t}'\}_{t=1}^T$, and $\bar{s}>0$ satisfy \eqref{eq:phi_reg} with 
\[
\phi=\phi_t,\quad \phi'=\phi_t',\quad s=\begin{cases}
\bar{s}, & t=1\bmod\ell,\\
0, & \text{otherwise},
\end{cases}
\quad \forall t\in[T].
\]
Moreover, given $Y_{0},Y_{0}'\in\dom\psi$, let $Z_{t}\sim{\cal N}(0,\sigma^{2}I)$,
and define the random variables
\[
\begin{gathered}Y_{t}:=\prox_{\psi}(\phi_{t}(Y_{t-1})+Z_{t}),\quad Y_{t}':=\prox_{\psi}(\phi_{t}'(Y_{t-1}')+Z_{t}), \quad \forall t\geq 1.\end{gathered}
\]
If $T=\ell$, then the following statements hold:
\begin{itemize}
    \item[(a)]if $\zeta=0$, then
    \begin{equation}
D_{\alpha}(Y_{T}\|Y_{T}')-D_{\alpha}^{(\tau)}(Y_{0}\|Y_{0}')\leq\frac{\alpha}{2}\left(\frac{L\tau+\bar{s}}{\sigma}\right)^{2}\theta_{L}(T);\label{eq:div_bd1}
\end{equation}
    \item[(b)]if $\tau=0$, $L=1$, and $\zeta=\bar{s}$, then 
    \begin{equation}
        D_{\alpha}(Y_{T}\|Y_{T}')-D_{\alpha}(Y_{0}\|Y_{0}')\leq 2\alpha T\left(\frac{\zeta}{\sigma}\right)^2
    \end{equation}
\end{itemize}
\end{lemma}

\begin{proof}
(a) Let $s=\bar{s}$. Our goal is to recursively apply \eqref{eq:main_recurrence} with
suitable choices of the free parameter $a$ at each application.
Specifically, let $\{(b_{t},R_{t},S_{T})\}$ be as in \eqref{eq:tech_bt},
and define
\[
a_{t}:=(L\tau+s)b_{t}\quad\forall t\geq1.
\]
Using Lemma~\ref{lem:tech_bt}(a)--(b), we first have $L\tau+s-a_{1}=(L\tau+s)R_{1}\geq0$
and, hence, by Lemma~\ref{lem:main_recurrence}, we have
\[
D_{\alpha}^{([L\tau+s]R_{1})}(Y_{1}\|Y_{1}')=D_{\alpha}^{(L\tau+{s}-a_{1})}(Y_{1}\|Y_{1}')\leq D_{\alpha}^{(\tau)}(Y_{0}\|Y_{0}')+\frac{\alpha a_{1}^{2}}{2\sigma^{2}}.
\]
Since Lemma~\ref{lem:tech_bt}(a)--(b) also implies $R_{t}\geq0$
and we have $s_{t}=0$ for $t\geq2$, we repeatedly apply Lemma~\ref{lem:main_recurrence}
with $(a,\tau)=(a_{t},\tau_{t})=(a_{t},0)$ for $t\geq2$ to obtain
\begin{align*}
D_{\alpha}^{(\tau)}(Y_{0}\|Y_{0}') & \geq D_{\alpha}^{([L\tau+s]R_{1})}(Y_{1}\|Y_{1}')-\frac{\alpha a_{1}^{2}}{2\sigma^{2}}
\geq D_{\alpha}^{([L\tau+s]LR_{1}-a_{2})}(Y_{2}\|Y_{2}')-\frac{\alpha(a_{1}^{2}+a_{2}^{2})}{2\sigma^{2}}\\
 & =D_{\alpha}^{([L\tau+s]R_{2})}(Y_{2}\|Y_{2}')-\frac{\alpha(a_{1}^{2}+a_{2}^{2})}{2\sigma^{2}}\geq\cdots \\
 & \geq D_{\alpha}^{([L\tau+s]R_{T})}(Y_{T}\|Y_{T}')-\frac{\alpha\sum_{i=1}^{T}a_{i}^{2}}{2\sigma^{2}}  =D_{\alpha}^{(0)}(Y_{T}\|Y_{T}')-\frac{\alpha\sum_{i=1}^{T}a_{i}^{2}}{2\sigma^{2}} \\
 & =D_{\alpha}(Y_{T}\|Y_{T}')-\frac{\alpha\sum_{i=1}^{T}a_{i}^{2}}{2\sigma^{2}}.
\end{align*}
It now remains to bound $\alpha\sum_{i=1}^{T}a_{i}^{2}/(2\sigma^2)$.
Using Lemma~\pageref{lem:tech_bt}(c) and the fact that $T=\ell$ and $\bar{s}=s$,
we have 
\begin{align*}
\frac{\alpha\sum_{i=1}^{T}a_{i}^{2}}{2\sigma^{2}} & =\frac{\alpha}{2\sigma^{2}}\left[(L\tau+s)^{2}\sum_{i=2}^{T}b_{i}^{2}\right]\leq\frac{\alpha}{2}\left(\frac{L\tau+\bar{s}}{\sigma}\right)^{2}\theta_{L}(T).
\end{align*}
Combining this bound with the previous one yields the desired conclusion.

(b) Let $s=\bar{s}$. Similar to (a), our goal is to recursively apply \eqref{eq:main_recurrence} with
suitable choices of the free parameter $a$ at each application. For this setting, let $a_1=\zeta+s$ and $a_t=\zeta$ for $t\geq 2$. Using the fact that $\tau=0$ and $L=1$ and Lemma~\ref{lem:main_recurrence}, we first have that
\[
D_\alpha(Y_1\|Y_1')= D_\alpha^{(0)}(Y_1\|Y_1') = D_\alpha^{(s + \zeta - a_1)}(Y_1\|Y_1') \leq D_\alpha^{(0)}(Y_0\|Y_0') + \frac{\alpha a_1^2}{2\sigma^2}.
\]
We then repeatedly apply Lemma~\ref{lem:main_recurrence} with $(a,\tau)=(a_{t},0)$ for $t\geq2$ to obtain
\begin{align*}
D_{\alpha}^{(0)}(Y_{0}\|Y_{0}') & \geq D_{\alpha}^{(0)}(Y_{1}\|Y_{1}')-\frac{\alpha a_{1}^{2}}{2\sigma^{2}}
\geq D_{\alpha}^{(\zeta-a_{2})}(Y_{2}\|Y_{2}')-\frac{\alpha(a_{1}^{2}+a_{2}^{2})}{2\sigma^{2}}\\
 & =D_{\alpha}^{(0)}(Y_{2}\|Y_{2}')-\frac{\alpha(a_{1}^{2}+a_{2}^{2})}{2\sigma^{2}}\geq\cdots \\
 & \geq D_{\alpha}^{(\zeta - a_T)}(Y_{T}\|Y_{T}')-\frac{\alpha\sum_{i=1}^{T}a_{i}^{2}}{2\sigma^{2}}  =D_{\alpha}^{(0)}(Y_{T}\|Y_{T}')-\frac{\alpha\sum_{i=1}^{T}a_{i}^{2}}{2\sigma^{2}} \\
 & =D_{\alpha}(Y_{T}\|Y_{T}')-\frac{\alpha\sum_{i=1}^{T}a_{i}^{2}}{2\sigma^{2}}.
\end{align*}
It now remains to bound $\alpha\sum_{i=1}^{T}a_{i}^{2}/(2\sigma^2)$. Using the definition of $\{a_t\}$ and the fact that $\zeta=s$, it holds that
\[
\frac{\alpha\sum_{i=1}^{T}a_{i}^{2}}{2\sigma^{2}}
\leq\frac{\alpha}{2\sigma^{2}}\left[4\zeta^2 + (T-1)\zeta^2\right] \leq 
2\alpha T\left(\frac{\zeta}{\sigma}\right)^2.
\]
Combining this bound with the previous one yields the desired conclusion.
\end{proof}

We next extend the above result to multiple dataset passes. 
\begin{proposition}
\label{prop:tech_div}
Let $L$, $\tau$, $\zeta$, $\bar{s}$, $\{Y_t\}$, $\{Y_t'\}$, $\ell$, and $T$ be as in Lemma~\ref{lem:single_pass_div}.
Moreover, let $\theta_L(\cdot)$ be as in \eqref{eq:dh_theta_def}.
For any $\tau\geq0$ and $E=\lfloor T/\ell\rfloor$,
the following statements hold:
\begin{itemize}
\item[(a)] 
if $\zeta=0$, then
\begin{align}
 & D_{\alpha}(Y_{T}\|Y_{T}')-D_{\alpha}^{(\tau)}(Y_{0}\|Y_{0}')
 \nonumber \\ & \leq\frac{\alpha}{2\sigma^{2}}\left[(L\tau+\bar{s})^{2}\theta_{L}(\ell)+\bar{s}^{2}\left\{ (E-1)\theta_{L}(\ell)+\theta_{L}(T-E\ell)\right\} \right].\label{eq:div_bd2}
\end{align}
\item[(b)]
if $\tau=0$ and $\zeta = \bar{s}$, then 
\begin{align}
 & D_{\alpha}(Y_{T}\|Y_{T}')-D_{\alpha}(Y_{0}\|Y_{0}') \leq 2\alpha T\left(\frac{\zeta}{\sigma}\right)^2.
 \label{eq:div_bd3}
\end{align}
\end{itemize}
\end{proposition}
\begin{proof}
(a) Let $s=\bar{s}$. For convenience, define
\begin{align*}
{\cal B}_{1}(\tau,T) & :=\frac{\alpha}{2}\left(\frac{L\tau+s}{\sigma}\right)^{2}\theta_{L}(T), \quad
{\cal B}_{2} :=\frac{\alpha}{\sigma^{2}}\left[(L\tau+s)^{2}+s^{2}\left\{ (E-1)\theta_{L}(\ell)+\theta_{L}(T-E\ell)\right\} \right].
\end{align*}
Using Lemma~\ref{lem:single_pass_div}(a), we have that for the first $\ell$
iterates,
\[
D_{\alpha}(Y_{\ell}\|Y_{\ell}')-D_{\alpha}^{(\tau)}(Y_{0}\|Y_{0}')\leq{\cal B}_{1}(\tau,\ell).
\]
Similarly, using part Lemma~\ref{lem:single_pass_div}(a) with $\tau=0$, we have that
\[
D_{\alpha}(Y_{[k+1]\ell}\|Y_{[k+1]\ell}')-D_{\alpha}^{(0)}(Y_{\ell}\|Y_{\ell}')\leq{\cal B}_{1}(0,\ell),
\]
for any $1\leq k\leq E-1$. Finally, using part Lemma~\ref{lem:single_pass_div}(a) with $T=T-E\ell$
and $\tau=0$, we have that
\[
D_{\alpha}(Y_{T}\|Y_{T}')-D_{\alpha}^{(0)}(Y_{E\ell}\|Y_{E\ell}')\leq{\cal B}_{1}(0,T-E\ell).
\]
Summing the above three inequalities, using the fact that $D_{\alpha}^{(0)}(X\|Y)=D_{\alpha}(X\|Y)$,
and using the definition of ${\cal B}_{2}$ we conclude that 
\begin{align*}
 & D_{\alpha}(Y_{T}\|Y_{T}')-D_{\alpha}(Y_{0}\|Y_{0}')
 \leq{\cal B}_{1}(\tau,\ell)+(E-1){\cal B}_{1}(0,\ell)+{\cal B}_{1}(0,T-E\ell)={\cal B}_{2}.
\end{align*}

(b) The proof follows similarly to (a). Repeatedly using Lemma~\ref{lem:single_pass_div}(b) at increments of $\ell$ iterations up to iteration $E\ell$, we have that
\begin{align*}
D_{\alpha}(Y_{E\ell}\|Y_{E\ell}')& \leq D_{\alpha}(Y_{(E-1)\ell}\|Y_{(E-1)\ell}')+ 2\alpha \ell\left(\frac{\zeta}{\sigma}\right)^2
\leq D_{\alpha}(Y_{(E-2)\ell}\|Y_{(E-2)\ell}')+4\alpha \ell\left(\frac{\zeta}{\sigma}\right)^2
\leq \cdots \\
& \leq D_{\alpha}(Y_{0}\|Y_{0}')+2\alpha E\ell\left(\frac{\zeta}{\sigma}\right)^2.
\end{align*}
For the last $T - E\ell$ iterations, we use Lemma~\ref{lem:single_pass_div}(b) with $T=T-E\ell$ and the above bound to obtain
\begin{align*}
D_{\alpha}(Y_{T}\|Y_{T}') &\leq D_{\alpha}(Y_{E\ell}\|Y_{E\ell}')+
2\alpha [T-E\ell]\left(\frac{\zeta}{\sigma}\right)^2 
\leq D_{\alpha}(Y_{0}\|Y_{0}')+2\alpha T\left(\frac{\zeta}{\sigma}\right)^2.
\end{align*}
\end{proof}
Some remarks about Proposition~\ref{prop:tech_div} are in order. First, part (a) shows that if $\phi_t$ and $\phi_t'$ only differ at $t=1$, then $D_\alpha(Y_T\|Y_T')$ is finite for any $T$. Second, part (a) also shows that if $\phi_t$ and $\phi_t'$ differ cyclically for a cycle length of $\ell$, then the divergence between $Y_T$ and $Y_T'$ grows linearly with the number of cycles $E$. Third, part (b) gives a bound that is independent of $L$. Finally, both of the bounds in parts (a) and (b) can be viewed as  R\'enyi divergences between the Gaussian random variables ${\cal N}(0, \sigma^2I)$ and ${\cal N}(\mu, \sigma^2I)$ for different values of $\mu$.

In Appendix~\ref{app:residuals}, we give a detailed discussion of how the residuals $a$ from Lemma~\ref{lem:main_recurrence} are chosen to prove Proposition~\ref{prop:tech_div}(a). In particular, we prove that the chosen residuals yield the tightest RDP bound that can achieved by repeatedly applying \eqref{eq:main_recurrence}.

\subsection{SGD operator analysis}
\label{app:sgd_analysis}

This subappendix derives some important properties about the DP-SGD update operator $\cal{A}_\lam(\cdot)$ in \eqref{eq:gen_dp_sgd_update} and also contains the proof of Proposition~\ref{prop:prox_lipsch}.

To start, we recall the following well-known bound from convex analysis. Its proof can be found, for example, in \cite[Theorem 5.8(iv)]{beck2017first}.
\begin{lemma}
\label{lem:co-coercive}
Let $F:\dom h\mapsto\r$ be convex and differentiable. Then $F$ satisfies
\begin{equation}
F(x)-F(y)-\left\langle \nabla F(y),x-y\right\rangle \leq\frac{L}{2}\|x-y\|^{2}\quad\forall x,y\in\dom h\label{eq:phi_descent}
\end{equation}
 if and only if 
\[
\left\langle \nabla F(x)-\nabla F(y),x-y\right\rangle \geq\frac{1}{L}\|\nabla F(x)-\nabla F(y)\|^{2}\quad\forall x,y\in\dom h.
\]
\end{lemma}
We next give a technical bound on $f_{i}$, which generalizes the
co-coercivity of convex functions to weakly-convex functions.
\begin{lemma}
\label{lem:tech_nonconv_cc}
For any $x,y\in\dom h$ and $f_i$ satisfying (A2), it holds that 
\begin{align*}
 & \left\langle \nabla f_{i}(x)-\nabla f_{i}(y),x-y\right\rangle  \geq-m\left[1+\frac{m}{2(M+m)}\right]\|x-y\|^{2}+\frac{1}{2(M+m)}\|\nabla f_{i}(x)-\nabla f_{i}(y)\|^{2}.
\end{align*}
\end{lemma}
\begin{proof}
Define $F=f_{i}+m\|\cdot\|^{2}/2$ and let $x,y\in\dom h$ be fixed.
Moreover, note that $F$ is convex and satisfies \eqref{eq:phi_descent}
with $L=M+m$. It then follows from Lemma~\ref{lem:co-coercive}
with $L=M+m$ that
\begin{align*}
\frac{1}{M+m}\|\nabla F(x)-\nabla F(y)\|^{2} & =\frac{1}{M+m}\|\nabla f_{i}(x)-\nabla f_{i}(y)+m(x-y)\|^{2}\\
 & \leq\left\langle \nabla F(x)-\nabla F(y),x-y\right\rangle \\
 & =\left\langle \nabla f_{i}(x)-\nabla f_{i}(y),x-y\right\rangle +m\|x-y\|^{2}.
\end{align*}
Applying the bound $\|a+b\|^{2}/2\leq\|a\|^{2}+\|b\|^{2}$ with $a=\nabla f_{i}(x)-\nabla f_{i}(y)+m(x-y)$
and $b=-m(x-y)$, the above inequality then implies 
\begin{align*}
\left\langle \nabla f_{i}(x)-\nabla f_{i}(y),x-y\right\rangle  & \geq-m\|x-y\|^{2}+\frac{1}{M+m}\|\nabla f_{i}(x)-\nabla f_{i}(y)+m(x-y)\|^{2}\\
 & \geq-\left[m+\frac{m^{2}}{2(M+m)}\right]\|x-y\|^{2}+\frac{1}{2(M+m)}\|\nabla f_{i}(x)-\nabla f_{i}(y)\|^{2}.
\end{align*}
\end{proof}
The below result gives some technical bounds on changes in the proximal function.

\begin{lemma}
\label{lem:prox} Given
$u,v\in\rn$, let $\psi\in\cConv(\rn)$ and define
\[
\Delta:=u-v,\quad\Delta^{p}:=\prox_{\psi}(x)-\prox_{\psi}(y).
\]
Then, the following statements hold:
\begin{itemize}
\item[(a)] $\|\Delta^{p}\|^2\leq\left\langle \Delta,\Delta^{p}\right\rangle $;
\item[(b)] $\|\Delta^{p}-\Delta\|^{2}\leq\|\Delta\|^{2}-\|\Delta^{p}\|^{2}$.
\end{itemize}
\end{lemma}
\begin{proof}
(a) See \cite[Theorem 6.42(a)]{beck2017first}.

(b) Using part (a), we have that
\begin{align*}
\|\Delta^{p}-\Delta\|^{2} & =\|\Delta^{p}\|^2+\|\Delta\|^{2}-2\left\langle \Delta,\Delta^{p}\right\rangle \leq\|\Delta\|^{2}-\|\Delta^{p}\|^{2}. 
\end{align*}
\end{proof}

We now develop some technical properties of ${\cal A}_{\lam}(\cdot)$.  The first result presents a bound involving the following quantities for $x,y\in\dom h$ and $i\in[k]$.
\begin{equation}
    \begin{gathered}d:=x-y,\quad
\Delta_{i}:=\nabla f_{i}(x)-\nabla f_{i}(y),\quad\Delta_{i}^{p}:=\prox_{q_{i}}(\nabla f_{i}(x))-\prox_{q_{i}}(\nabla f_{i}(y)).
\end{gathered} \label{eq:deltas_def}
\end{equation}

\begin{lemma}
\label{lem:deltas_bd}
Let $x,y\in \dom h$ and $i\in[k]$ be fixed, let $d$, $\Delta_i $, and $\Delta_i^p$ be as in \eqref{eq:deltas_def} for some $\{f_i\}$. Moreover, let $L_\lam$ be as in \eqref{eq:L_lam},
and suppose $f_i$ satisfies assumption (A2). If $\Delta_i^p = \Delta_i$, then for any $\lam\leq 1/(M+m)$ we have
\begin{equation}
\|d - \lam \Delta_i^p\| \leq L_\lam \|d\|. \label{eq:L_lam_bd}
\end{equation}
On the other hand, if $\Delta_i^p \neq \Delta_i$, then for any $\lam\leq 1/[2(M+m)]$ we have
\begin{equation}
\|d - \lam \Delta_i^p\| \leq \sqrt{2} L_\lam \|d\|. \label{eq:sqrt2_L_lam_bd}
\end{equation}
\end{lemma}
 
\begin{proof}
Before proceeding, we first establish a technical inequality. Using Lemma~\ref{lem:tech_nonconv_cc}, it holds that for any $\mu >0$, 
\begin{align}
 \mu \left\Vert \Delta_{i}\right\Vert ^{2}-2\left\langle d,\Delta_{i}\right\rangle 
 &\leq \mu \|\Delta_{i}\|^{2} + 2m\left[1+\frac{m}{2(M+m)}\right]\|d\|^{2}-\frac{\|\Delta_{i}\|^{2}}{M+m}\nonumber \\
 & = 2m\left[1+\frac{m}{2(M+m)}\right]\|d\|^{2}+\left({\mu -\frac{1}{M+m}}\right)\|\Delta_{i}\|^{2}.\label{eq:tech_lipsch2}
\end{align}
We now prove \eqref{eq:L_lam_bd}. Supposing that $\Delta_i^p = \Delta_i$, we have 
\[
\|d - \lam \Delta_i^p\|^2 = \|d - \lam \Delta_i\|^2 = \|d\|^2 + \lam\left[\lam\|\Delta_i\|^2 - 2 \left\langle d,\Delta_{i}\right\rangle\right].
\]
Using \eqref{eq:tech_lipsch2} with $\mu = \lam$, the above identity, and the definition of $L_\lam$, it holds that for any 
$\lam\leq1/(M+m)$, we have 
\begin{align*}
 \|d - \lam \Delta_i^p\|^2 & \leq \left(1+ 2\lam m\left[1+\frac{m}{2(M+m)}\right]\right) \|d\|^2 + \lam\left({\lam -\frac{1}{M+m}}\right)\|\Delta_{i}\|^{2} \\
 & = L_\lam^2 \|d\|^2 + \lam\left({\lam -\frac{1}{M+m}}\right)\|\Delta_{i}\|^{2} \leq L_\lam^2 \|d\|^2.
\end{align*}

We now prove \eqref{eq:sqrt2_L_lam_bd}. Using Lemma~\ref{lem:prox}(b) with $(\Delta, \Delta^p)=(\Delta_i, \Delta_i^p)$ and the inequality $\|a+b\|^2 \leq 2\|a\|^2 + 2\|b\|^2$ for $a,b\in\rn$, it holds that 
\begin{align*}
\left\Vert d-\lam\Delta_{i}^{p}\right\Vert ^{2} & 
= \left \Vert d-\lam(\Delta_{i}+\Delta_{i}-\Delta_{i}^{p})\right\Vert ^{2}
\leq 2\|d-\lam\Delta_{i}\|^{2} + 2\lam^2 \|\Delta_{i}-\Delta_{i}^{p}\|^{2} \\
 & \overset{Lem.~\ref{lem:prox}(b)}{\leq} 2 \|d-\lam\Delta_{i}\|^{2}+2\lam^{2}\|\Delta_{i}\|^{2}-2\lam^{2}\|\Delta_{i}^p\|^{2}\\
 & \leq 2 \|d-2\lam\Delta_{i}\|^{2}+2\lam^{2}\|\Delta_{i}\|^{2} \\
 & = 2\left(\|d\|^{2}+
 \lam\left[2\lam\left\Vert \Delta_{i}\right\Vert ^{2}- 2\left\langle d,\Delta_{i}\right\rangle\right]\right).
\end{align*}
Using \eqref{eq:tech_lipsch2} with $\mu = 2\lam$, the above inequality, and the definition of $L_\lam$, it holds that for any 
$\lam\leq1/[2(M+m)]$, we have 
\begin{align*}
 \|d - \lam \Delta_i^p\|^2 
 & \leq 2\left(1 + 2\lam m\left[1+\frac{m}{2(M+m)}\right]\right) \|d\|^2 + 2\left({2\lam -\frac{1}{M+m}}\right)\|\Delta_{i}\|^{2} \\
 & = 2 L_\lam^2 \|d\|^2 + 2\left({2\lam -\frac{1}{M+m}}\right)\|\Delta_{i}\|^{2} \leq 2 L_\lam^2 \|d\|^2. 
\end{align*}
\end{proof}

We are now ready to give the proof of Proposition~\ref{prop:prox_lipsch}.

\subsubsection{Proof of Proposition~\ref{prop:prox_lipsch}}

\begin{proof}
(a) Let $x,y\in\dom h$ and $\lam > 0$ be arbitrary. Moreover, denote $p_{i}(\cdot)=\prox_{q_{i}}(\cdot)$ for $i\in[k]$. Using
the definition of ${\cal A}_{\lam}(\cdot)$, the assumption that $\|p_{i}(z)\|\leq C$
for any $z\in\rn$, and the triangle inequality, we have that
\begin{align*}
\|{\cal A}_{\lam}(x)-{\cal A}_{\lam}(y)\| & =\left\Vert x-y+\frac{\lam}{|B|}\sum_{i\in B}[p_{i}(x)-p_{i}(y)]\right\Vert \leq\|x-y\|+\frac{\lam}{|B|}\sum_{i\in B}(\|p_{i}(x)\|+\|p_{i}(y)\|)\\
 & \leq\|x-y\|+2\lam C.
\end{align*}

(b) Let $x,y\in\dom h$ be arbitrary, and denote
$\xi(\cdot):={\cal A}_{\lam}(\cdot,\{f_{i}\},\{q_{i}\})$.
Moreover, let $d$, $\Delta_i$, and $\Delta_i^p$ be as in \eqref{eq:deltas_def}.
 Using the fact that $\|\sum_{i=1}^{|B|}v_{i}\|^{2}\leq |B|\sum_{i=1}^{|B|}\|v_{i}\|^{2}$
for any $\{v_{i}\}\subseteq\rn$, we have
\begin{align}
\|\xi(x)-\xi(y)\|^{2} & =\frac{1}{|B|^{2}}\left\Vert \sum_{i\in B}\left\{ \left[x-\lam\prox_{q_{i}}(\nabla f_{i}(x))\right]-\left[y-\lam\prox_{q_{i}}(\nabla f_{i}(y))\right]\right\} \right\Vert ^{2}\nonumber \\
 & =\frac{1}{|B|^{2}}\left\Vert \sum_{i\in B}(d-\lam\Delta_{i}^{p})\right\Vert ^{2}\leq\frac{1}{|B|}\sum_{i\in B}\left\Vert d-\lam\Delta_{i}^{p}\right\Vert ^{2}. \label{eq:tech_lipsch1}
\end{align}
Using \eqref{eq:tech_lipsch1} and \eqref{eq:sqrt2_L_lam_bd} in Lemma~\ref{lem:deltas_bd}, we conclude that
\begin{align*}
\|\xi(x)-\xi(y)\|^{2} & \leq \frac{1}{|B|}\sum_{i\in B}\left\Vert d-\lam\Delta_{i}^{p}\right\Vert ^{2} \leq 2 L_\lam^2 \|d\|^2 = 2 L_\lam \|x - y\|^2.
\end{align*}

(c) Let $\xi(\cdot)$, $d$, $\Delta_i$, and $\Delta_i^p$ be as in part (b). Following the same argument as in part (b), we obtain \eqref{eq:tech_lipsch1}. Using \eqref{eq:tech_lipsch1} and \eqref{eq:L_lam_bd} in Lemma~\ref{lem:deltas_bd}, we conclude that
\begin{align*}
\|\xi(x)-\xi(y)\|^{2} & \leq \frac{1}{|B|}\sum_{i\in B}\left\Vert d-\lam\Delta_{i}^{p}\right\Vert ^{2} \leq L_\lam^2 \|d\|^2 = L_\lam \|x - y\|^2.
\end{align*}
\end{proof}

\subsection{RDP bounds}

This subappendix derives the RDP bounds in Theorems~\ref{thm:dp_sgd_bd1} and \ref{thm:general_compl}.

The first result shows how the updates in Algorithm~\ref{alg:dp_sgd} are  randomized proximal updates applied to the operator $A_\lam(\cdot)$ in \eqref{eq:gen_dp_sgd_update} with $q_i(\cdot) = {\rm Clip}_C(\cdot)$.

\begin{lemma}
\label{lem:dp_sgd_equiv1}
Let $\{X_{t}\}$, $\{X_{t}'\}$, $\lam$, $b$, $\sigma$,
$C$, and $T$ be as in Theorem~\ref{thm:dp_sgd_bd1}. Moreover, denote
\[
\phi_{t}(x):={\cal A}_{\lam}(x,\{f_{i}\},\{{\rm Clip}_C\}),\quad\phi_{t}'(x):={\cal A}_{\lam}(x,\{f_{i}'\},\{{\rm Clip}_C\}),\quad\forall x\in\dom h,
\]
where ${\rm Clip}_C(\cdot)$ and ${\cal A}_\lam(\cdot)$ are as in \eqref{eq:ClipC_def} and \eqref{eq:gen_dp_sgd_update}, respectively. Then, it holds that
\[
X_{t}=\prox_{\lam h}(\phi_{t}(X_{t-1})+N_{t}),\quad X_{t}'=\prox_{\lam h}(\phi_{t}'(X_{t-1}')+N_{t}),\quad\forall t\geq1.
\]
\end{lemma}
\begin{proof}
This follows immediately from the definition
of $\phi_{t}$, the update rules in Algorithm~\ref{alg:dp_sgd}, and the fact that ${\rm Clip}_C(\cdot)$ is the proximal operator of the (convex) indicator function of the convex set $\{x: \|x\|\leq C\}$.
\end{proof}

We now present some important norm bounds.

\begin{lemma}
\label{lem:displacement}
Let $\{X_{t}\}$, $\{X_{t}'\}$, $\{\phi_t\}$, and $\{\phi_t'\}$
be as in Theorem~\ref{thm:dp_sgd_bd1} and denote $\ell = k/b$ and $t^{*}:=\inf_{t\geq1}\left\{ t:i^{*}\in B_{t}\right\}$.
Then, it holds that
\begin{equation}
\|X_{t^{*}}-X_{t^{*}}'\|=0,\quad\|\phi_{s}(x)-\phi_{s}'(x)\|\leq\frac{2\lam C}{b},\label{eq:xs_bd}
\end{equation}
for every $s\in\{j\ell+t^{*}:j=0,1,...\}$ and any $x\in\dom h$.
\end{lemma}

\begin{proof}
The identity in \eqref{eq:xs_bd} follows from the fact that $X_{t}=X_{t}'$
for every $t\leq t^{*}$. For the inequality in \eqref{eq:xs_bd},
it suffices to show the bound for $s=t^{*}$ because the batches $B_{t}$
in Algorithm\ \ref{alg:dp_sgd} are drawn cyclically. To that end,
let $x\in\dom h$ be fixed. Using the update rule in Algorithm\ \ref{alg:dp_sgd},
and the fact that $\|{\rm Clip}_{C}(x)\|\leq C$ for every $x\in\rn$,
we have that 
\begin{align*}
\|\phi_{s}(x)-\phi_{s}'(x)\| & =\frac{1}{b}\left\Vert \sum_{i\in B_{t^{*}}}\left[x-\lam{\rm Clip}_{C}(\nabla f_{i^{*}}(x))\right]-\left[x-\lam{\rm Clip}_{C}(\nabla f_{i^{*}}'(x))\right]\right\Vert \\
 & =\frac{\lam}{b}\|{\rm Clip}_{C}(\nabla f_{i^{*}}(x))-{\rm Clip}_{C}(\nabla f_{i^{*}}'(x))\|\\
 & \leq\frac{\lam}{b} \left[\|{\rm Clip}_{C}(\nabla f_{i^{*}}(x))\|+\|{\rm Clip}_{C}(\nabla f_{i^{*}}'(x))\|\right]=\frac{2\lam C}{b}.
\end{align*}
\end{proof}

We now give the proofs of the main RDP bounds.

\subsubsection{Proof of Theorem~\ref{thm:dp_sgd_bd1}}

\begin{proof}
% [Proof of Theorem~\ref{thm:dp_sgd_bd1}]
Using Proposition~\ref{prop:tech_div}(a) with
\[
Y_{0}=X_{T-1},\quad Y_{0}'=X_{T-1}',\quad\tau=d_{h},\quad s=\frac{2\lam C}{b},\quad L=L_{\lam},\quad\ell=\frac{k}{b},
\]
and $E=T=1$, we have that 
\begin{align*}
D_{\alpha}(X_{T}\|X_{T}') & \leq D_{\alpha}^{(d_{h})}(X_{T-1}\|X_{T-1}')+\frac{\alpha}{2\sigma^{2}}\left(Ld_{h}+\frac{2\lam C}{b}\right)^{2}\theta_{L}(1)=\frac{\alpha}{2\sigma^{2}}\left(Ld_{h}+\frac{2\lam C}{b}\right)^{2}.
\end{align*}
\end{proof}

\subsubsection{Proof of Theorem~\ref{thm:general_compl}}

\begin{proof}
Suppose $f_{i^*}' \neq f_{i^*}$ and $i^* \in B_{t^*}$ for indices $i^*$ and $t^*$.
We first prove \eqref{eq:dp_sgd_indep_bd1}. 
In view of Proposition~\ref{prop:prox_lipsch}(b), it is clear that the DP-SGD update is $\sqrt{2} L_\lam$-Lipschitz continuous. Hence,
using Lemma~\ref{lem:displacement}(b) and Proposition~\ref{prop:tech_div}(a) with
\[
Y_{0}=X_{t^{*}},\quad Y_{0}'=X_{t^{*}}',\quad\tau=0,\quad s=\frac{2\lam C}{b},\quad L=L_{\lam},\quad\ell=\frac{k}{b},
\]
and $T=T-t^{*}-1$, we have that
\begin{align*}
D_{\alpha}(X_{T}\|X_{T}') & \leq D_{\alpha}^{(0)}(X_{0}\|X_{0})+2\alpha\left(\frac{\lam C}{b\sigma}\right)^{2}\left[E\theta_{L_{\lam}}(\ell)+\theta_{L_{\lam}}(T-t^{*}-1-E\ell)\right]\\
 & =2\alpha\left(\frac{\lam C}{b\sigma}\right)^{2}\left[E\theta_{L_{\lam}}(\ell)+\theta_{L_{\lam}}(T-t^{*}-1-E\ell)\right] \leq 2\alpha\left(\frac{\lam C}{b\sigma}\right)^{2}\left[1 + E\theta_{L_{\lam}}(\ell)\right], 
\end{align*}
where the last inequality follows from the fact that $\theta_{L_\lam}(s)$ is nonincreasing for $s\geq 1$ and $\theta_{L_\lam}(1) = 1$.

We now prove \eqref{eq:dp_sgd_indep_bd2}. In view of \eqref{eq:noclip_cond} and Proposition~\ref{prop:prox_lipsch}(c), it is clear that the DP-SGD update, in the absence of gradient clipping, is $L_\lam$-Lipschitz continuous. Consequently, the desired bound follows from the same arguments as in the proof of \eqref{eq:dp_sgd_indep_bd2} above, but with $L=L_\lam$ instead of $L=\sqrt{2}L_\lam$. 
\end{proof}

\section{Choice of residuals}

\label{app:residuals}

This appendix briefly discusses the choice of residuals $\{a_t\}$ that
are used in the proof of Proposition~\ref{prop:tech_div}(a) and
Lemma~\ref{lem:main_recurrence}. 

In the setup of Proposition~\ref{prop:tech_div}(a), it is straightforward
to show that if $\{a_{t}\}$ is a nonnegative sequence of scalars
such that 
\[
\tilde{R}_{t}:=L^{t-1}(L\tau+s)-\sum_{i=1}^{t}a_{i}L^{t-i}\geq0,\quad\tilde{R}_{T}=0,
\]
then repeatedly applying Lemma~\ref{lem:main_recurrence} with $a=a_{t}$
yields
\begin{equation}
D_{\alpha}(Y_{T}\|Y_{T}')-D_{\alpha}^{(\tau)}(Y_{0}\|Y_{0}')\leq\frac{\alpha}{2\sigma^{2}}\sum_{i=1}^{T}a_{i}^{2}.\label{eq:gen_div_bd}
\end{equation}
Hence, to obtain the tightest bound of the form in \eqref{eq:gen_div_bd},
we need to solve the quadratic program 
\begin{align*}
(P)\quad\min\  & \frac{1}{2}\sum_{i=1}^{T}a_{i}^{2}\\
\text{s.t}\  & \tilde{R}_{t}\geq0\quad\forall t\in[T-1],\\
 & \tilde{R}_{T}=0.
\end{align*}
If we ignore the inequality constraints, the first order optimality
condition of the resulting problem is that there exists $\xi\in\r$
such that 
\[
a_{i}=\xi L^{t-i}\quad\forall t\in[T],\quad\tilde{R}_{T}=0.
\]
The latter identity implies that 
\[
L^{T-1}(L\tau+s)=\xi\sum_{i=1}^{T}L^{2(T-i)}=\xi\sum_{i=0}^{T-1}L^{2i}
\]
which then implies 
\[
a_{i}=\frac{L^{T-1}(L\tau+s)L^{t-i}}{\sum_{i=0}^{T-1}L^{2i}}\quad\forall t\in[T].
\]
Hence, to verify that the above choice of $a_{i}$ is optimal for $(P)$,
it remains to verify that $\tilde{R}_{t}\geq0$ for $t\in[T-1]$.
Indeed, this follows from Lemma~\ref{lem:tech_bt}(b) after normalizing
for the $L\tau+s$ factor. As a consequence, the right-hand-side of
\eqref{eq:gen_div_bd} is minimized for our choice of $a_{i}$ above.

\section{Parameter choices}
\label{app:params}
Let us now consider some interesting values for $\lam$, $\sigma$, and $\ell$.

The result below establishes a useful bound on $\theta_L(s)$ for sufficiently large enough values of $s$.

\begin{lemma}
\label{lem:theta_bd_upper}
For any $L>1$ and $\xi>1$, if $s\geq\log_{L}\sqrt{\xi/(\xi-1)}$
then $\theta_{L}(s)\leq\xi\left(1-L^{-2}\right).$
\end{lemma}
\begin{proof}
Using the definition of $\theta_{L}(\cdot)$, we have 
\begin{align*}
\theta_{L}(s) & =\frac{L^{-2(s-1)}}{\sum_{i=0}^{s-1}L^{2i}}=\frac{L^{2s}-L^{2(s-1)}}{L^{2s}-1}=\frac{1-L^{-2}}{1-L^{-2s}}\leq\frac{1-L^{-2}}{1-L^{-2\log_{L}\sqrt{\xi/(\xi-1)}}}\\
 & =\frac{1-L^{-2}}{1-(\xi-1)/\xi}=\xi\left(1-\frac{1}{L^{2}}\right). 
\end{align*}
\end{proof}
\begin{corollary}
\label{cor:spec_bd}
Let $\alpha>1$ and $\varepsilon>0$ be fixed,
and let $\{X_{t}\}$, $\{X_{t}'\}$, $b$, $C$, $E$, $\ell$, $\lam$,
and $T$ be as in Theorem~\ref{thm:general_compl}. Moreover, define
\[
\overline{\lam}(\rho):=\frac{1}{2(M+\rho)},\quad\overline{\sigma}_{\varepsilon}(\rho):=\frac{C\cdot\overline{\lam}(\rho)}{2b}\sqrt{\frac{1}{\alpha\varepsilon}\left(1+\left[\frac{4\rho}{M+\rho}\right]E\right)},\quad\overline{\ell}(\rho):=\frac{\log2}{\log\left[1+\rho\overline{\lam}(\rho)\right]},
\]
for every $\rho>0$. If $\lam=\overline{\lam}(m)$, $\sigma\geq\overline{\sigma}_{\varepsilon}(m)$,
 $\ell\geq\overline{\ell}(m)$, and no gradient clipping is performed, then 
\begin{align*}
D_{\alpha}(X_{T}\|X_{T}')  \leq4\alpha\left[\frac{C \cdot \overline{\lam}(m) }{b\cdot\overline{\sigma}_\varepsilon(m)}\right]^{2}\left[1+\frac{4m}{M+m}\right],
\end{align*}
and the corresponding instance of Algorithm~\ref{alg:dp_sgd} is $(\alpha,\varepsilon)$-R\'enyi-DP.
\end{corollary}
\begin{proof}
For ease of notation, denote $\overline{\lam}=\overline{\lam}(m)$, $L=L_{\overline{\lam}}$,
$\overline{\sigma}=\overline{\sigma}(m)$, and $\overline{\ell}=\overline{\ell}(m)$.
We first note that
\[
L=L_{\overline{\lam}}=\sqrt{1+\frac{m}{M+m}\left[1+\frac{m}{M+m}\right]}\geq\sqrt{1+m\overline{\lam}(m)},
\]
which implies 
\[
\ell\geq\overline{\ell}=\frac{\log\sqrt{2}}{\log\sqrt{1+m\overline{\lam}}}=\frac{\log\sqrt{2}}{\log L}=\log_{L}\sqrt{2}.
\]
Consequently, using Lemma~\ref{lem:theta_bd_upper} with $(\xi,s)=(2,\ell)$
and the definitions of $L_{\lam}(\cdot)$ and $\overline{\lam}(\cdot)$,
we have that 
\[
\theta_{L}(\ell)\leq2\left(1-\frac{1}{L^{2}}\right)=2\left(\frac{2m}{2(M+m)}\left[1+\frac{m}{M+m}\right]\right)\leq\frac{4m}{M+m}.
\]
Using the above bound and Theorem~\ref{thm:general_compl} with $(\lam,\sigma,L)=(\overline{\lam},\overline{\sigma},L_{\overline{\lam}})$,
we obtain
\begin{align*}
D_{\alpha}(X_{T}\|X_{T}') & \leq4\alpha\left(\frac{\overline{\lam}C}{b\sigma}\right)^{2}\left[1+E\theta_{L}(\ell)\right]\leq4\alpha\left(\frac{\overline{\lam}C}{b\sigma}\right)^{2}\left[1+\frac{4Em}{M+m}\right]\\
 & \leq4\alpha\left(\frac{\overline{\lam}C}{b\overline{\sigma}}\right)^{2}\left[1+\frac{4Em}{M+m}\right]\leq\varepsilon.
\end{align*}
In view of the fact that Algorithm~\ref{alg:dp_sgd} returns the
last iterate $X_{T}$ (or $X_{T}'$), the conclusion follows.
\end{proof}

Some remarks about Corollary~\ref{cor:spec_bd} are in order. First, $\sigma_{\varepsilon}^2(m)$ increases linearly with the number of dataset passes $E$. Second, the smaller $m$ is the smaller the effect of $E$ on $\sigma_{\varepsilon}(m)$ is. 
Fourth, $\lim_{m\to 0} \overline{\ell}(m)=\infty$ which implies that the reducing the dependence on $E$ in $\sigma_{\varepsilon}(m)$ leads to more restrictive choices on $\ell$.
Finally, it is worth emphasizing that the restrictions on $\ell$ can be removed by using \eqref{eq:dp_sgd_indep_bd1} directly. However, the resulting bounds are less informative in terms of the topological constants $m$ and $M$.

We now present an RDP bound that is independent of $E$ when $\lam$ is sufficiently small.

\begin{corollary}
Let $\{X_{t}\}$, $\{X_{t}'\}$, $b$, $C$, $E$, $\ell$, $\lam$, $\sigma$,
and $T$ be as in Theorem~\ref{thm:general_compl}. If \[
\lam\leq\min\left\{\frac{1}{\sqrt{E}},\frac{1}{2(m+M)}\right\}
\]
and no gradient clipping is performed, then we have
\[
D_{\alpha}(X_{T}\|X_{T}')\leq4\alpha\left(\frac{C}{b\sigma}\right)^{2}\left[1+\theta_{L_{\lam}}(\ell)\right].
\]
\end{corollary}
\begin{proof}
Using Theorem~\ref{thm:general_compl} and the fact that $\theta_{L}(\cdot)\leq1$
for any $L>1$, we have that 
\begin{align*}
D_{\alpha}(X_{T}\|X_{T}') & \leq4\alpha\left(\frac{\lam C}{b\sigma}\right)^{2}\left[\theta_{L_{\lam}}(T-E\ell)+E\theta_{L_{\lam}}(\ell)\right]\leq4\alpha\left(\frac{\lam C}{b\sigma}\right)^{2}\left[1+E\theta_{L_{\lam}}(\ell)\right]\\
 & \leq4\alpha\left(\frac{C}{b\sigma}\right)^{2}\left[\frac{1}{E}+\theta_{L_{\lam}}(\ell)\right]\leq4\alpha\left(\frac{C}{b\sigma}\right)^{2}\left[1+\theta_{L_{\lam}}(\ell)\right]. 
\end{align*}
\end{proof}

\section{Limitations of Poisson sampling in practice}
\label{app:random_batch_sampling}

This appendix discussing the computational limitation of implementing Poisson sampling in practice. It is primarily concerned with the large-scale setting where datasets may be on the order of hundreds of millions of examples.

\vspace*{1em}
\noindent\textit{Data access}. Implementations of Poisson sampling, e.g., Opacus \cite{opacus}, typically employ a pseudorandom number generator to (i) randomly sample a collection of indices from zero to $N-1$, where $N$ is the size of the dataset and (ii) map these indices to corresponding examples in the dataset to generate a batch of examples. In order for (ii) to be efficient, many libraries need fast random access to the dataset which is difficult to do without loading the entire dataset into RAM (as reading data from disk can be orders of magnitude more expensive). In contrast, cyclic traversal of batches only requires (relatively small) fixed blocks of the dataset to be loaded into memory for every batch and need not perform a matching of indices (such as in (i) above) to data.

\vspace*{1em}
\noindent\textit{Variable batch size}. Independent of the access speed of the dataset examples, Poisson sampling also generates batches of random sizes, which are typically inconvenient to handle in deep learning systems \cite{CGK24}. For example, popular just-in-time compilation-based machine learning libraries such as JAX, PyTorch/Opacus, and TensorFlow may need to retrace their computation graph at every training step as the batch size cannot be statically inferred or kept constant. 
Additionally, optimizing training workloads on hardware accelerators such as graphical processing units (GPUs) and tensor processing units (TPUs) becomes difficult as (i) they require any in-device data to have fixed sizes and (ii) any input data generated by Poisson sampling will have variable sizes due to the effect of variable batch sizes. In contrast, the cyclic traversal of batches will always generate fixed batch sizes and, consequently, will not suffer from the above issues.

\end{document}